\documentclass{article}

\PassOptionsToPackage{numbers,compress}{natbib}
\usepackage[main,final]{neurips_2026}

\usepackage[utf8]{inputenc} % allow utf-8 input
\usepackage[T1]{fontenc}    % use 8-bit T1 fonts
\usepackage{url}            % simple URL typesetting
\usepackage{booktabs}       % professional-quality tables
\usepackage{amsfonts}       % blackboard math symbols
\usepackage{nicefrac}       % compact symbols for 1/2, etc.
\usepackage{microtype}      % microtypography
\usepackage{xcolor}         % colors

\usepackage{amsmath}
\usepackage{enumitem}
\usepackage{algorithm}
\usepackage{algpseudocode}
\usepackage{graphicx}
\usepackage{pifont}
\usepackage{float}
\usepackage{wrapfig}
\usepackage{colortbl,xcolor}   % \rowcolor
\definecolor{rowgray}{gray}{0.92}   % ours 的灰底
\usepackage{multirow}
\usepackage{hyperref}
\title{\textsc{MicroWorld}: Empowering Multimodal Large Language Models to Bridge the Microscopic Domain Gap with Multimodal Attributed Property Graph}

\author{%
  Manyu Li \quad Ruian He \quad  Chenxi Ma\thanks{Corresponding authors.} \quad Weimin Tan\footnotemark[1] \quad  Bo Yan\footnotemark[1] \\
  College of Computer Science and Artificial Intelligence \\
  Shanghai Key Laboratory of Intelligent Information Processing \\
  Fudan University, Shanghai, China \\
  \texttt{manyuli24@m.fudan.edu.cn} \\
  \texttt{\{rahe16,cxma,wmtan,byan\}@fudan.edu.cn}
}

\begin{document}

\maketitle

\begin{abstract}
Multimodal large language models (MLLMs) show remarkable potential for scientific reasoning, yet their performance in specialized domains such as microscopy remains limited by the scarcity of domain-specific training data and the difficulty of encoding fine-grained expert knowledge into model parameters.
To bridge the gap, we introduce \textsc{MicroWorld}, a framework for structured-knowledge augmentation with a complementary visual retrieval pathway, using a \emph{multimodal attributed property graph} (MAPG) without domain-specific fine-tuning.
\textsc{MicroWorld} extracts biomedical entities and relations via scispaCy or LLM-based triplet mining, aligns images and entities in a shared embedding space using Qwen3-VL-Embedding, and assembles a knowledge graph comprising approximately 106K nodes and 2.62M typed edges spanning eight relation categories.
At inference time, a graph-augmented retrieval pipeline matches query entities to the MAPG and injects structured knowledge context into the MLLM prompt.
On MicroVQA, \textsc{MicroWorld} raises Qwen3-VL-8B-Instruct accuracy from 48.8\% to \textbf{67.1\%} (+18.3 percentage points; 37.5\% relative), exceeding \emph{zero-shot} GPT-5 (59.4\%). The identical retrieved context raises GPT-5 to \textbf{71.5\%}, and GPT-5's own entity extraction reaches 73.3\%. On MicroBench, 8B Single-Pass improves overall accuracy by \textbf{4.0 percentage points}. The benefits are greatest for knowledge-intensive reasoning; fine-grained visual discrimination and the additional visual-pathway contribution remain limited. Code and accompanying resources: \href{https://github.com/ieellee/MicroWorld}{https://github.com/ieellee/MicroWorld}.

\end{abstract}

\section{Introduction}

The rapid advancement of artificial intelligence is fundamentally reshaping how we understand the natural world and conduct biomedical research~\cite{merchant2023scaling,hoogeboom2022equivariant,cui2024scgpt}.
Foundation models have driven breakthroughs across a wide spectrum of scientific domains, from predicting protein structures with atomic accuracy~\cite{abramson2024accurate} to accelerating drug discovery~\cite{jumper2021highly} and enabling automated disease diagnosis~\cite{rajpurkar2022ai}.
In microscopy, modern imaging modalities such as cryo-electron microscopy, super-resolution fluorescence, and high-throughput confocal imaging are generating data at an unprecedented scale and resolution~\cite{weigert2018content,ouyang2018deep}, far outpacing the capacity of human experts to analyze, interpret, and derive actionable insights.
This widening gap between data acquisition and scientific understanding calls for intelligent systems that can reason about microscopy images at the level of domain experts, assisting in experimental result interpretation, hypothesis generation, and experimental design, and ultimately participate in a \emph{lab-in-the-loop} paradigm that tightly integrates AI reasoning into the scientific discovery cycle~\cite{boiko2023autonomous,szymanski2023autonomous}.

However, building domain-specialized MLLMs for microscopy faces a fundamental tension.
Fine-tuning a foundation model on a narrow scientific sub-field demands massive volumes of high-quality, expert-annotated instruction data, which is prohibitively expensive and labor-intensive to curate in microscopy, where each annotation requires deep domain expertise~\cite{li2023llava,zhang2024multimodal}.
Moreover, models fine-tuned on limited domain data often suffer from poor generalization and catastrophic forgetting of broader capabilities~\cite{kotha2023understanding,luo2025empirical}.
An alternative paradigm is retrieval-augmented generation (RAG), which enriches the input context with relevant external knowledge at inference time, circumventing the need for extensive domain-specific training~\cite{lewis2020retrieval,yu2024rankrag,zakka2024almanac}.
Recent studies have reinforced this insight: for specialized scientific domains with scarce instruction-tuning data, augmenting the reasoning context of general-purpose MLLMs can be more effective and robust than fine-tuning~\cite{zhang2025agentic,gao2023retrieval}, as it preserves the model's broad reasoning capabilities while injecting domain-specific knowledge precisely where it is needed. This raises a natural question: \textit{\textbf{can we leverage a scientific literature database to construct a multimodal knowledge base that enhances the reasoning capabilities of MLLMs in the microscopy domain without supervised fine-tuning?}}

\begin{figure}[t]
  \centering
  \includegraphics[width=1\textwidth]{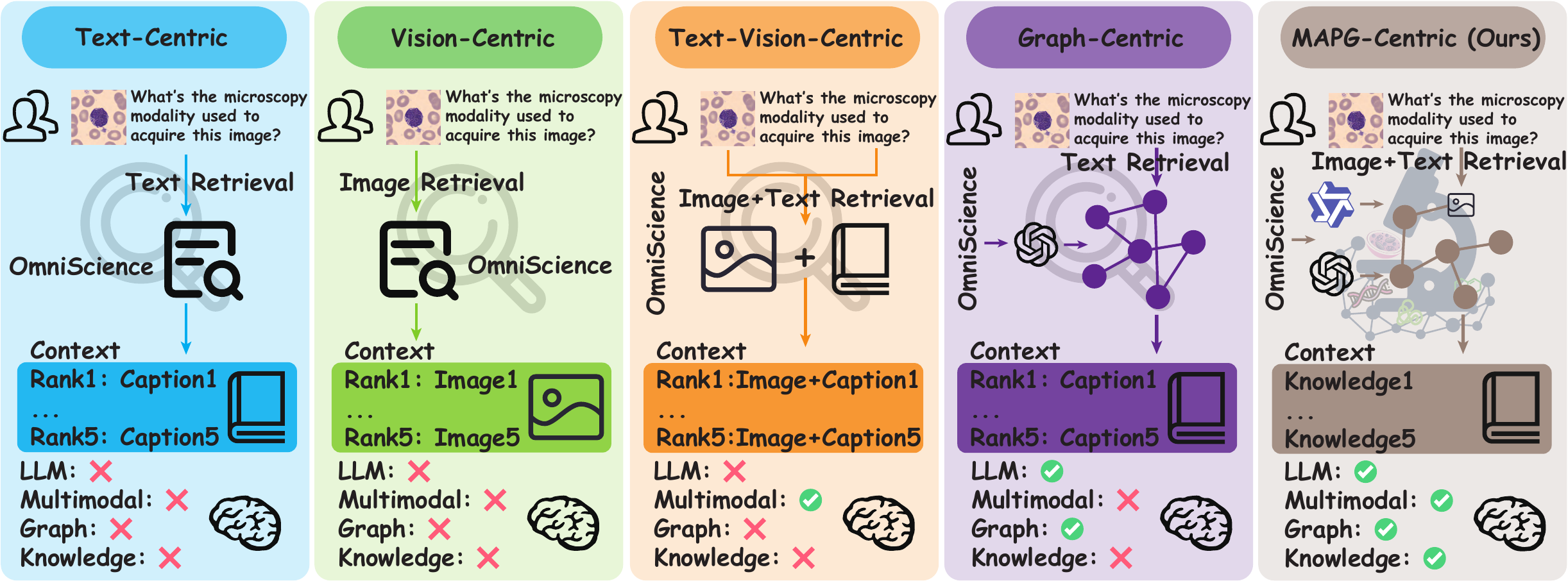}
  \caption{Comparison with prior RAG paradigms. While \textbf{Text-Centric} , \textbf{Vision-Centric} , and \textbf{Text-Vision-Centric} methods rely respectively on captions, images, or a multimodal combination of both, \textbf{Graph-Centric} RAG builds a caption-based structured graph but lacks domain-specific modeling. Conversely, our \textbf{MAPG-Centric} approach is tailored for biological microscopy and naturally integrates visual modalities. \textbf{LLM}: Utilization of a Large Language Model. \textbf{Multimodal}: Use of multimodal features during retrieval. \textbf{Graph}: Construction of a structured retrieval space. \textbf{Knowledge}: Provision of specialized biological microscopy knowledge.}
  \label{fig:related}
\end{figure}

A key enabler for this vision is the emergence of large-scale scientific corpora.
OmniScience~\cite{tao2026omniscience} has assembled over 1.5 million image--caption pairs sourced from top-tier journals (with an average impact factor exceeding 12) and high-citation preprint servers (arXiv, bioRxiv, medRxiv), and further employed MLLMs to reconstruct high-quality, self-contained captions.
Among this vast collection, approximately 692K pairs are related to microscopy, encompassing rich interactions among domain-specific entities: biological structures, imaging techniques, staining protocols, morphological properties, and organism types.
These densely interconnected scientific descriptions present a unique opportunity to construct a comprehensive, structured knowledge base for microscopy that goes beyond flat image-caption pair retrieval to capture the relational fabric of the domain.

In this paper, we introduce \textsc{MicroWorld}, a framework that bridges the microscopic domain gap for MLLMs by constructing and leveraging a multimodal attributed property graph (MAPG) tailored for microscopy to introduce domain knowledge.
Our main contributions are as follows:

\begin{itemize}[leftmargin=1.5em, itemsep=3pt]
  \item \textbf{A large-scale multimodal knowledge graph for microscopy.}
  We construct \textsc{MicroWorld}, a multimodal attributed property graph comprising approximately 106K entity and image nodes interconnected by 2.62M typed edges spanning eight relation categories.
  By integrating biomedical entity extraction, LLM-based relation mining, and vision--language alignment via Qwen3-VL-Embedding, \textsc{MicroWorld} captures both the semantic structure and visual characteristics of the microscopy domain within a unified graph representation.

  \item \textbf{Knowledge-augmented microscopy reasoning without supervised fine-tuning.}
  \textsc{MicroWorld} raises Qwen3-VL-8B-Instruct from 48.8\% to 67.1\% on MicroVQA, above zero-shot GPT-5 (59.4\%). Giving GPT-5 the identical context reaches 71.5\%, demonstrating that the knowledge also benefits a stronger backbone.

  \item \textbf{A dynamic, extensible knowledge infrastructure.}
  \textsc{MicroWorld} is designed as a living knowledge system that supports incremental graph augmentation from heterogeneous scientific corpora.
  Its modular architecture allows seamless integration of new data sources, enabling the knowledge graph to evolve alongside the rapidly advancing microscopy literature and to be adapted to diverse experimental contexts without retraining.
\end{itemize}

\section{Related work}

\subsection{Multimodal large language models in microscopy}

Recent advances in multimodal large language models (MLLMs) have demonstrated remarkable emergent capabilities in understanding and reasoning over complex data, showing immense potential as intelligent assistants for accelerating biological discoveries~\cite{zhang2024multimodal}. To facilitate this progress, OmniScience~\cite{tao2026omniscience} curated a large-scale dataset of images and captions from premier journals and conferences in biological sciences, and reconstructed self-contained captions to improve data quality. However, in the microscopy sub-domain of biological sciences, the scarcity of large-scale, high-quality annotated data makes it extremely challenging to fine-tune a dedicated microscopy foundation model~\cite{zhang2024multimodal}. To evaluate MLLM capabilities in this domain, several benchmarks have been proposed: MicroBench~\cite{lozano2024micro} focuses on fundamental perception-level tasks, while MicroVQA~\cite{burgess2025microvqa} targets more advanced scientific reasoning abilities. In this work, we construct \textsc{MicroWorld}, a comprehensive framework that substantially enhances MLLM reasoning performance on microscopy tasks.

\subsection{Retrieval-augmented generation in the biomedical field}

Retrieval-augmented generation (RAG) has been extensively explored for natural images, with representative approaches spanning text-centric methods such as FLMR~\cite{lin2023fineflmr}, vision-centric methods including VisRAG~\cite{yu2024visrag} and ColPali~\cite{faysse2024colpali}, as well as text-vision-centric frameworks like REVEAL~\cite{hu2023reveal}. In the biomedical domain, MasonNLP~\cite{karim2025masonnlp} leverages a generalist, instruction-tuned large language model coupled with a RAG framework to integrate in-domain textual and visual exemplars, demonstrating the benefits of retrieval-based knowledge grounding. MMed-RAG~\cite{xia2024mmed} further introduces a versatile multimodal RAG system specifically designed to enhance the factual accuracy of MLLMs in medical contexts. Following the development of MMed-RAG, RAG methodologies for medical VQA have continuously evolved and improved~\cite{wu2025mkgf, deria2026medmo}. Despite these advances, the application of RAG in microscopy remains largely unexplored. Our proposed \textsc{MicroWorld} bridges this gap by introducing a retrieval-augmented pipeline tailored for microscopy, offering new insights for future research in this specialized domain.

\subsection{Graph-structured knowledge in multimodal large language models}

Graph-structured representations offer a natural and expressive means of encoding relational knowledge, capturing latent associations among entities that flat image-caption pair stores or unstructured corpora cannot easily reveal~\cite{woo2018linknet,li2022graph,mo2025kggen}. GraphRAG~\cite{edge2024localgraphrag} pioneers a graph-based approach to question answering over private text corpora, constructing a community-level summarization of an LLM-generated knowledge graph to handle queries that require holistic reasoning across the entire dataset, scaling effectively with both the generality of user questions and the volume of source text. LightRAG~\cite{guo2024lightrag} further advances this direction by incorporating graph structures into text indexing and retrieval, employing a dual-level retrieval paradigm that enables comprehensive information discovery at both fine-grained entity-level and coarse-grained topic-level granularities. Inspired by these developments, we propose \textsc{MicroWorld}, which carefully models the rich relational structure inherent in microscopy data, including associations among specimens, imaging modalities, staining techniques, and morphological features to construct a multimodal attributed property graph, thereby uncovering latent connections that can guide more informed and accurate scientific reasoning. Figure~\ref{fig:related} shows the comparison between our method and prior RAG paradigms.

\section{Method}
\label{sec:method}

We present \textsc{MicroWorld}, a framework that constructs a \emph{multimodal attributed property graph} (MAPG) from scientific image--caption corpora and leverages it to augment MLLM reasoning on microscopy VQA.
As shown in Figure~\ref{fig:method}, the pipeline consists of three stages:
(i)~\emph{triplet extraction and multimodal embedding} (\S\ref{sec:stage1}),
(ii)~\emph{graph construction} (\S\ref{sec:stage2}),
and (iii)~\emph{knowledge-grounded inference} (\S\ref{sec:stage3}).

\paragraph{Notation.}
Let $\mathcal{D} = \{(x_i, c_i)\}_{i=1}^{N}$ be $N$ image--caption pairs.
We define the MAPG as $\mathcal{G} = (\mathcal{V},\, \mathcal{E},\, \mathcal{R},\, \boldsymbol{\phi},\, \boldsymbol{\psi})$,
where $\mathcal{V} = \mathcal{V}_e \cup \mathcal{V}_x$ contains entity nodes~$\mathcal{V}_e$ and image nodes~$\mathcal{V}_x$;
$\mathcal{E} \subseteq \mathcal{V} \times \mathcal{R} \times \mathcal{V}$ are typed edges over relation vocabulary~$\mathcal{R}$;
$\boldsymbol{\phi}: \mathcal{V} \to \mathbb{R}^d$ maps nodes to embeddings;
and $\boldsymbol{\psi}: \mathcal{V}_e \to \mathcal{A}$ assigns structured attributes.
Each entity node has one of five semantic types: \textsc{Structure}, \textsc{Method}, \textsc{Condition}, \textsc{Property}, or \textsc{Organism}; image nodes belong to the separate set~$\mathcal{V}_x$.
Given a VQA instance $(x_q, q, \{a_k\}_{k=1}^{K})$, the goal is to predict~$k^{*}$ by conditioning on subgraph knowledge from~$\mathcal{G}$.

\subsection{Stage~1: Triplet extraction and multimodal embedding}
\label{sec:stage1}

This stage transforms the corpus~$\mathcal{D}$ (filtered from OmniScience via the method in Appendix~\ref{app:corpus_selection}) into structured triplets and dense embeddings.

\begin{figure}[t]
\centering
\includegraphics[width=1\textwidth]{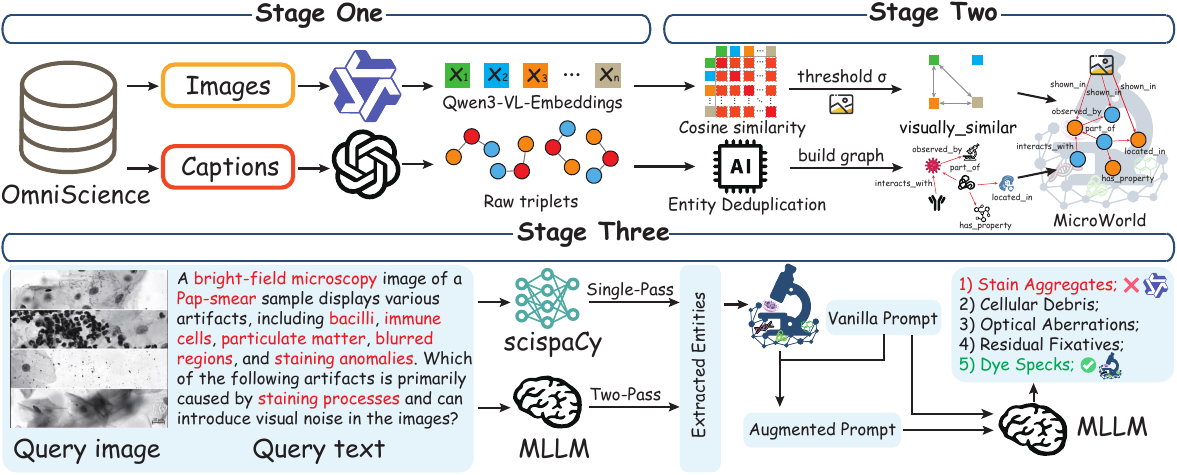}
\caption{Overview of the \textsc{MicroWorld} framework. \textbf{Stage~1}: Images and captions from OmniScience are processed via scispaCy NER and LLM-based relation extraction to produce raw triplets, while Qwen3-VL-Embedding computes dense vision--language representations. \textbf{Stage~2}: Extracted entities undergo deduplication and are assembled into a multimodal attributed property graph (MAPG) with eight relation types; image embeddings are compared via cosine similarity to add \texttt{visually\_similar} edges above threshold~$\sigma$. \textbf{Stage~3}: At inference time, entities are extracted from the query via scispaCy (Single-Pass) or the MLLM itself (Two-Pass), matched to the MAPG, and the retrieved knowledge context is injected into an augmented prompt for the final answer.}
\label{fig:method}
\end{figure}

\paragraph{Entity and relation extraction.}
For each caption~$c_i$, we apply \texttt{scispaCy}~\citep{neumann2019scispacy} NER with noun-chunk augmentation to extract biomedical entities, each linked to MeSH identifiers and classified into one of the five semantic types via heuristic rules (details in Appendix~\ref{app:method_details}).
We then prompt GPT-5~\cite{singh2025openai} to extract relation triplets $\{(h, r, t)\}$ constrained to five biologically meaningful types:
\texttt{observed\_by}, \texttt{has\_property}, \texttt{located\_in}, \texttt{interacts\_with}, and \texttt{part\_of} (see Figure~\ref{fig:method}).
Additionally, for each caption with $m$ entities, we add $\binom{m}{2}$ undirected \texttt{co\_occurs\_with} edges:
\begin{equation}
  \mathcal{E}_{\text{co}} = \bigl\{(e_j,\, \texttt{co\_occurs\_with},\, e_k) \mid 1 \le j < k \le m,\ e_j, e_k \in \text{Entities}(c_i)\bigr\}.
  \label{eq:cooccur}
\end{equation}

\paragraph{Vision--language embedding.}
We employ Qwen3-VL-Embedding~\citep{li2026qwen3} to compute $\ell_2$-normalized $d$-dimensional embeddings for images ($\mathbf{h}_x$), image--caption pairs ($\mathbf{h}_m$), and entity names ($\mathbf{h}_e$), using task-specific instructions.
These embeddings are used in Stage~2 for visual similarity edges and in Stage~3 for retrieval.

\subsection{Stage~2: Graph construction}
\label{sec:stage2}

This stage assembles the extracted triplets and embeddings into the MAPG~$\mathcal{G}$.

\paragraph{Entity deduplication and graph assembly.}
An entity registry merges surface forms with identical canonical names (after lowercasing, punctuation removal, and whitespace collapsing), unifying aliases, MeSH IDs, and provenance.
For each image~$x_i$, we create an image node $v_{x_i} \in \mathcal{V}_x$ connected to its caption entities via \texttt{shown\_in} edges:
$\mathcal{E}_{\text{vis}} = \bigl\{(v_{x_i},\, \texttt{shown\_in},\, v_e) \mid v_e \in \text{Entities}(c_i)\bigr\}$.
The full edge set $\mathcal{E} = \mathcal{E}_{\text{bio}} \cup \mathcal{E}_{\text{co}} \cup \mathcal{E}_{\text{vis}} \cup \mathcal{E}_{\text{vsim}}$ uses bidirectional adjacency for non-\texttt{shown\_in} relations.

\paragraph{Entity description enrichment.}
We query NCBI Entrez Utilities~\cite{sayers2021database} for authoritative definitions with type-dependent lookup strategies, falling back to GPT-5-generated definitions when NCBI results are absent or too short ($<$40 chars).

\paragraph{Visual similarity edges.}
\label{sec:vis_sim_edges}
We augment~$\mathcal{G}$ with \texttt{visually\_similar} edges between image pairs whose embedding cosine similarity exceeds~$\sigma = 0.85$:
\begin{equation}
  \mathcal{E}_{\text{vsim}} = \bigl\{(v_{x_i},\, \texttt{visually\_similar},\, v_{x_j}) \mid \boldsymbol{\phi}(v_{x_i})^{\top}\boldsymbol{\phi}(v_{x_j}) \ge \sigma,\; i < j\bigr\}.
  \label{eq:vis_sim}
\end{equation}

\paragraph{Structural precomputation.}
To support efficient online retrieval, we precompute (i)~adaptive $k$-hop neighborhoods (expanding to 2-hop for low-degree nodes, capping at 50 neighbors for hubs) and (ii)~a hybrid similarity score $S(v_i, v_j) = \alpha \cdot J(v_i, v_j) + (1 - \alpha) \cdot C(v_i, v_j)$ that fuses Jaccard structural similarity with cosine embedding similarity (default $\alpha = 0.5$).
Full formulations are in Appendix~\ref{app:method_details}.
The graph also supports incremental augmentation from external corpora (Appendix~\ref{app:graph_augment}).
Table~\ref{tab:relations} in the Appendix summarizes all eight relation types.

\subsection{Stage~3: Knowledge-grounded inference}
\label{sec:stage3}

At inference time, given a query $(x_q, q)$, we retrieve relevant subgraph context and inject it into the MLLM prompt (Algorithm~\ref{alg:retrieval} in the Appendix).

\paragraph{Single-pass inference.}
\label{sec:single_pass}
ScispaCy NER extracts entity mentions from~$q$, which are matched to the KG via cascaded exact-alias and fuzzy matching ($\rho_{\text{fuzzy}} = 0.72$); overly generic entities (frequency $> \rho_{\text{skip}}$) are filtered.
We encode~$x_q$ with the same Qwen3-VL-Embedding encoder and retrieve nearest image nodes among the 20K stored vectors, then expand over \texttt{visually\_similar} and \texttt{shown\_in} edges; up to three visual entities supplement the text matches. The query need not be a graph node.
For each matched entity, we assemble a knowledge block with its definition, semantic type, and $k$-hop neighbors, then prepend the assembled context~$\mathcal{C}$ to the question.

\paragraph{Two-pass inference.}
\label{sec:two_pass}
The MLLM first extracts up to 8 domain-specific entities from~$(x_q, q)$ (Round~1), which are then matched against the KG.
A fresh single-turn prompt with the retrieved knowledge is used for the final answer (Round~2), avoiding multi-turn context degradation.

\section{Experiments}
\label{sec:experiments}

\subsection{Experimental setup}
\label{sec:exp_setup}

\paragraph{Dataset.}
We evaluate on MicroVQA and MicroBench. MicroVQA is a microscopy visual question answering benchmark containing 1{,}042 multiple-choice questions spanning three scientific reasoning tasks:
\emph{Hypothesis Generation} (420 samples), \emph{Perception} (392 samples), and \emph{Experiment Proposal} (230 samples). MicroBench is a microscopy visual question answering benchmark that contains 17{,}315 images of multiple-choice questions that focus on \emph{Perception}.

\paragraph{Backbone models.}
We evaluate \textsc{MicroWorld} with two families of multimodal large language models at three scales:
\textbf{Qwen3-VL}~\citep{bai2025qwen3} (2B, 4B, 8B) and \textbf{InternVL3.5}~\cite{wang2025internvl3} (2B, 4B, 8B).

\paragraph{Inference modes.}
We compare two \textsc{MicroWorld} inference paradigms:
\emph{Single-Pass}, where scispaCy extracts entities from the question before the MLLM forward pass (\S\ref{sec:single_pass}), and
\emph{Two-Pass}, where the MLLM first extracts entities, followed by KG retrieval and a single-turn knowledge-augmented answer (\S\ref{sec:two_pass}).

\paragraph{RAG baselines.}
We compare against six representative RAG methods spanning different modality focuses:
\emph{text-centric}: FLMR~\citep{lin2023fineflmr};
\emph{vision-centric}: VisRAG~\citep{yu2024visrag} and ColPali~\citep{faysse2024colpali};
\emph{vision-text}: REVEAL~\citep{hu2023reveal};
\emph{graph-centric}: GraphRAG~\citep{edge2024localgraphrag} and LightRAG~\citep{guo2024lightrag}.
All baselines use the same filtered 20K OmniScience corpus, backbone, prompt template, and decoding configuration. We sweep top-$k \in \{1,3,5\}$ and report each method's best setting. Indexing and retrieved content differ by method: MAPG additionally supplies entity definitions and typed relations, so shared source data do not imply identical knowledge content.

\paragraph{Metrics.}
We report per-task accuracy and overall accuracy of MicroVQA and MicroBench samples. In Appendix~\ref{app:latency}, we also report 95th-percentile (P95) latency and throughput (queries per second, QPS) for both single-pass and two-pass inference.

\begin{table}[t]
  \centering
  \caption{MicroVQA accuracy (\%). ``Baseline'' is zero-shot without KG context. GPT-5 ``shared'' uses identical Qwen3-VL-8B Two-Pass context; ``own R1'' uses GPT-5 entity extraction. GPT-5 overall scores are independently reported from task scores. InternVL3.5 uses frozen Qwen3-VL prompts. Best commercial and per-size open-source results are \textbf{bolded}. \textsuperscript{*}Results from MicroVQA leaderboard.}
  \label{tab:model_comparison}

  \setlength{\tabcolsep}{3pt}
\begin{tabular}{llcccc}
  \toprule
  \textbf{Model} & \textbf{Mode} & \textbf{Hyp.~Gen.} & \textbf{Perception} & \textbf{Exp.~Prop.} & \textbf{Overall} \\
  \midrule
  \multicolumn{6}{l}{\textit{Commercial Models (zero-shot / MAPG-augmented)}} \\
  o1\textsuperscript{*}~\cite{openai_o1_series} & Baseline & 50.2 & 53.0 & 55.4 & 52.8 \\
  Claude Sonnet 4.5\textsuperscript{*}~\cite{anthropic_claude_sonnet_4_5_system_card_2025} & Baseline    & 56.4 & 49.6 & 55.1 & 54.4 \\
  o4-mini\textsuperscript{*}~\cite{openai_o3_o4mini_systemcard_2025} & Baseline & 56.1 & 50.4 & 57.9 & 55.6 \\
  o3\textsuperscript{*}~\cite{openai_o3_o4mini_systemcard_2025} & Baseline & 60.5 & 53.5 & 61.5 & 59.3 \\
  GPT-5\textsuperscript{*}~\cite{singh2025openai} & Baseline & 58.9 & 53.9 & 63.3 & 59.4 \\
  \rowcolor{rowgray}
  GPT-5 & MAPG shared & 72.5 & 71.0 & 70.4 & 71.5 \\
  \rowcolor{rowgray}
  GPT-5 & MAPG own & \textbf{74.3} & \textbf{73.2} & \textbf{71.7} & \textbf{73.3} \\
  \midrule
  \multicolumn{6}{l}{\textit{2B scale}} \\
  InternVL3.5 & Baseline    & 42.9 & 43.4 & 49.6 & 44.5 \\
  \rowcolor{rowgray}
  InternVL3.5 & Single & 49.5 & 49.2 & 58.3 & 51.3 \\
  \rowcolor{rowgray}
  InternVL3.5 & Two    & 53.3 & 55.4 & \textbf{63.5} & 56.3 \\
  Qwen3-VL    & Baseline    & 42.4 & 44.9 & 53.9 & 45.9 \\
  \rowcolor{rowgray}
  Qwen3-VL    & Single & 51.9 & 51.8 & 56.1 & 52.8 \\
  \rowcolor{rowgray}
  Qwen3-VL    & Two    & \textbf{60.0} & \textbf{57.4} & 61.7 & \textbf{59.4} \\
  \midrule
  \multicolumn{6}{l}{\textit{4B scale}} \\
  InternVL3.5 & Baseline    & 43.1 & 45.4 & 50.4 & 45.6 \\
  \rowcolor{rowgray}
  InternVL3.5 & Single & 43.8 & 45.9 & 43.0 & 44.4 \\
  \rowcolor{rowgray}
  InternVL3.5 & Two    & 48.1 & 51.0 & 50.9 & 49.8 \\
  Qwen3-VL    & Baseline    & 46.4 & 47.4 & 44.8 & 46.4 \\
  \rowcolor{rowgray}
  Qwen3-VL    & Single & 55.2 & 55.1 & 53.0 & 54.7 \\
  \rowcolor{rowgray}
  Qwen3-VL    & Two    & \textbf{62.9} & \textbf{66.6} & \textbf{63.0} & \textbf{64.3} \\
  \midrule
  \multicolumn{6}{l}{\textit{8B scale}} \\
  InternVL3.5 & Baseline    & 38.1 & 42.6 & 47.4 & 41.8 \\
  \rowcolor{rowgray}
  InternVL3.5 & Single & 55.0 & 55.6 & 53.0 & 54.8 \\
  \rowcolor{rowgray}
  InternVL3.5 & Two    & 64.0 & 63.5 & 62.2 & 63.4 \\
  Qwen3-VL    & Baseline    & 48.6 & 49.2 & 48.3 & 48.8 \\
  \rowcolor{rowgray}
  Qwen3-VL    & Single & 58.3 & 60.5 & 57.4 & 58.9 \\
  \rowcolor{rowgray}
  Qwen3-VL    & Two    & \textbf{67.9} & \textbf{68.4} & \textbf{63.5} & \textbf{67.1} \\
  \bottomrule
  \end{tabular}
  \end{table}

\paragraph{Implementation details.}
Local-model experiments use 4$\times$NVIDIA RTX 3090 GPUs with multi-GPU parallel inference.
Default hyperparameters are: $M_\text{text}=6$, $M_\text{vis}=3$, $\rho_\text{skip}=0.08$, $\rho_\text{compact}=0.04$, maximum context length 6{,}000 characters, $M_\text{nbr}=4$ neighbors, and a \texttt{max\_2hop} budget of 3.

\subsection{Main results}
\label{sec:main_results}

Table~\ref{tab:model_comparison} reports results across backbone models and inference modes.
\textbf{(1)~\textsc{MicroWorld} boosts zero-shot performance.}
Compared to the zero-shot baseline (no KG context), \textsc{MicroWorld} Two-Pass achieves large improvements across all models: Qwen3-VL-2B-Instruct improves from 45.9\% to 59.4\% (+13.5~pp), 4B from 46.4\% to 64.3\% (+17.9~pp), and 8B from 48.8\% to 67.1\% (+18.3~pp).
InternVL3.5 shows similar gains, with 8B jumping from 41.8\% to 63.4\% (+21.6~pp).
\textbf{(2)~Two-Pass consistently outperforms Single-Pass.}
Two-Pass yields 5.0--9.6~pp higher accuracy than Single-Pass across all six backbones; Single-Pass also gains +6.9 to +10.1~pp over baseline on Qwen3-VL.
\textbf{(3)~Backbone capability affects knowledge use.}
Qwen3-VL Two-Pass improves from 59.4\% (2B) through 64.3\% (4B) to 67.1\% (8B). InternVL3.5 is non-monotonic (56.3\%, 49.8\%, 63.4\%), so size alone does not determine performance.
\textbf{(4)~Cross-model transfer is effective.}
InternVL3.5-8B achieves 63.4\% using frozen prompts originally generated for Qwen3-VL, only 3.7~pp below the native Qwen3-VL-8B result (67.1\%).
This demonstrates that \textsc{MicroWorld}'s structured knowledge context is model-agnostic.
\textbf{(5)~GPT-5 also benefits from the same knowledge.}
The identical 8B Two-Pass context raises GPT-5 from 59.4\% to 71.5\% (+12.1~pp); its own Round-1 extraction reaches 73.3\%. Thus, the 67.1\% open-source result exceeds zero-shot GPT-5, while MAPG-augmented GPT-5 remains stronger.

\subsection{Comparison with RAG baselines}
\label{sec:rag_comparison}

Tables~\ref{tab:rag_comparison},~\ref{tab:rag_comparison_2b4b} show comparison results with RAG baselines (based on the same raw data as \textsc{MicroWorld}). \textsc{MicroWorld} outperforms all RAG baselines by a substantial margin across all three model scales.
On the 2B backbone, \textsc{MicroWorld} achieves 59.4\% overall, surpassing the best baseline (ColPali at 46.9\%) by \textbf{12.5~pp}.
On the 4B backbone, the gap widens to \textbf{16.6~pp} (64.3\% vs.\ GraphRAG at 47.7\%).
On the 8B backbone, \textsc{MicroWorld} reaches 67.1\%, exceeding the best baseline (FLMR at 48.7\%) by \textbf{18.4~pp}.
The advantage holds across all three task types, consistent with the value of entity-centric definitions and relation-linked knowledge. This comparison does not isolate graph structure from differences in knowledge content.
Among baselines, graph-centric methods (GraphRAG, LightRAG) generally outperform text-only (FLMR) and vision-only (VisRAG) at the 4B scale, while at 8B text-centric FLMR emerges as the strongest baseline, suggesting larger models extract more value from raw textual context.

\subsection{Ablation study}
\label{sec:ablation}

\begin{table}[t]
\centering
\caption{RAG comparison at 8B on MicroVQA: same filtered-20K corpus, backbone, prompt template, and decoding; best top-$k \in \{1,3,5\}$ per baseline. Type: T=text, V=vision, G=graph. \textsc{MicroWorld} uses Two-Pass and additional entity definitions and relations. 2B/4B results: Appendix~\ref{app:rag_2b4b}. Best results are \textbf{bolded}.}
\label{tab:rag_comparison}

\setlength{\tabcolsep}{4pt}
\begin{tabular}{llccccc}
\toprule
\textbf{Method} & \textbf{Type} & \textbf{$k$} & \textbf{Hyp.~Gen.} & \textbf{Perception} & \textbf{Exp.~Prop.} & \textbf{Overall} \\
\midrule
\rowcolor{rowgray}
\textsc{MicroWorld} (Ours) & G+V+T & --  & \textbf{67.9} & \textbf{68.4} & \textbf{63.5} & \textbf{67.1} \\
\midrule
FLMR              & T     & 1   & 48.6          & 50.0          & 46.5          & 48.7          \\
VisRAG            & V     & 1   & 47.1          & 50.3          & 47.8          & 48.5          \\
ColPali           & V     & 1   & 47.1          & 49.2          & 48.3          & 48.2          \\
GraphRAG          & G     & 1   & 46.9          & 51.8          & 46.1          & 48.6          \\
LightRAG          & G     & 3   & 45.5          & 48.7          & 47.8          & 47.2          \\
REVEAL            & T+V   & 3   & 46.2          & 45.9          & 50.0          & 46.9          \\
\bottomrule
\end{tabular}
\end{table}

\begin{table}[t]
\centering
\caption{MicroVQA Two-Pass ablations. $\Delta$ is the difference from default (pp). The joint 8B text-only control is reported separately in Appendix~\ref{app:knowledge_controls}.}
\label{tab:ablation}

\setlength{\tabcolsep}{3pt}
\begin{tabular}{llcccccc}
\toprule
 & & \multicolumn{2}{c}{\textbf{2B}} & \multicolumn{2}{c}{\textbf{4B}} & \multicolumn{2}{c}{\textbf{8B}} \\
\cmidrule(lr){3-4} \cmidrule(lr){5-6} \cmidrule(lr){7-8}
\textbf{Parameter} & \textbf{Value} & \textbf{Acc.} & \textbf{$\Delta$} & \textbf{Acc.} & \textbf{$\Delta$} & \textbf{Acc.} & \textbf{$\Delta$} \\
\midrule
Default & -- & 59.4 & -- & 64.3 & -- & 67.1 & -- \\
\midrule
\texttt{max\_text\_entities} & 3 & 59.9 & +0.5 & 62.5 & $-$1.8 & 64.6 & $-$2.5 \\
 & 10 & 59.7 & +0.3 & 64.6 & +0.3 & 66.1 & $-$1.0 \\
\midrule
\texttt{max\_context\_chars} & 4000 & 57.9 & $-$1.5 & 61.3 & $-$3.0 & 64.5 & $-$2.6 \\
 & 8000 & 59.6 & +0.2 & 66.1 & +1.8 & \textbf{68.7} & \textbf{+1.6} \\
\midrule
\texttt{max\_neighbors} & 3 & 59.3 & $-$0.1 & 64.3 & +0.0 & 67.7 & +0.6 \\
 & 6 & 59.0 & $-$0.4 & 64.3 & +0.0 & 67.6 & +0.5 \\
\midrule
\texttt{freq\_skip\_ratio} & 0.06 & 58.3 & $-$1.1 & 64.7 & +0.4 & 66.4 & $-$0.7 \\
 & 0.12 & 59.9 & +0.5 & 64.2 & $-$0.1 & 67.9 & +0.8 \\
\midrule
\texttt{freq\_compact\_ratio} & 0.02 & 59.4 & +0.0 & 65.5 & +1.2 & 67.9 & +0.8 \\
 & 0.08 & 60.0 & +0.6 & 64.5 & +0.2 & 66.5 & $-$0.6 \\
\midrule
\texttt{max\_visual\_entities} & 0 & 60.2 & +0.8 & 62.8 & $-$1.5 & 65.8 & $-$1.3 \\
 & 5 & 58.4 & $-$1.0 & 64.6 & +0.3 & \textbf{68.9} & \textbf{+1.8} \\
\midrule
\texttt{max\_2hop} & 0 & \textbf{62.0} & \textbf{+2.6} & 63.7 & $-$0.6 & 67.9 & +0.8 \\
 & 2 & 60.4 & +1.0 & 63.8 & $-$0.5 & 67.9 & +0.8 \\
\midrule
\texttt{visually\_similar} & - & 60.7 & +1.3 & 63.4 & $-$0.9 & 68.6 & +1.5 \\
\bottomrule
\end{tabular}
\end{table}

Table~\ref{tab:ablation} reports one-at-a-time (OAT) ablations on eight retrieval settings (Qwen3-VL, Two-Pass), with $\Delta$ relative to default. The joint 8B text-only control is analyzed separately below.
\textbf{(1)~Context length matters for larger models.}
Increasing \texttt{max\_context\_chars} from 6K to 8K improves 8B accuracy by 1.6~pp (to 68.7\%), while reducing it to 4K consistently hurts performance across all sizes ($-$1.5 to $-$3.0~pp).
Larger models can effectively utilize longer context windows.
\textbf{(2)~Visual entities are valuable at scale.}
Removing visual entities (\texttt{max\_visual\_entities}$=0$) reduces 4B and 8B accuracy by 1.5 and 1.3~pp, but improves 2B by 0.8~pp.
Increasing visual entities to 5 benefits the 8B model (+1.8~pp) but hurts the 2B model ($-$1.0~pp), suggesting that smaller models may struggle with additional visual context.
\textbf{(3)~2-hop expansion shows diminishing returns.}
Disabling 2-hop neighbors surprisingly improves the 2B model by 2.6~pp, while the effect on 4B and 8B is marginal ($<$1~pp).
This indicates that deeper graph traversal introduces noise for smaller models.
\textbf{(4)~Neighbor count is robust.}
Varying \texttt{max\_neighbors} between 3 and 6 has minimal impact ($\leq$0.6~pp), suggesting the default value of 4 is near-optimal.
\textbf{(5)~Visual similarity edges show mixed effects.}
Disabling \texttt{visually\_similar} edges ($\sigma = 0.85$, \S\ref{sec:vis_sim_edges}) yields +1.3~pp on 2B and +1.5~pp on 8B, but $-$0.9~pp on 4B.
The model-dependent effects suggest a trade-off between relevance, context length, and retrieval noise.

\paragraph{Joint and format controls.}
The 8B text-only KG reaches 66.0\% versus full MAPG's 67.1\% on MicroVQA, and 58.15\% versus 58.54\% on MicroBench (Two-Pass). These modest visual contributions are non-additive relative to the individual ablations. A separate 2-hop $\times$ visual-edge control yields a difference-in-differences of $-0.5$~pp (Appendix~\ref{app:factorial}). Rendering identical entities, definitions, and relations as a flat paragraph yields 66.9\% versus 67.1\% with structured blocks, suggesting limited sensitivity to presentation format (Appendix~\ref{app:knowledge_controls}).

\subsection{Knowledge graph scale study}
\label{sec:scale_study}

\begin{figure}[t]
\centering
\includegraphics[width=1\textwidth]{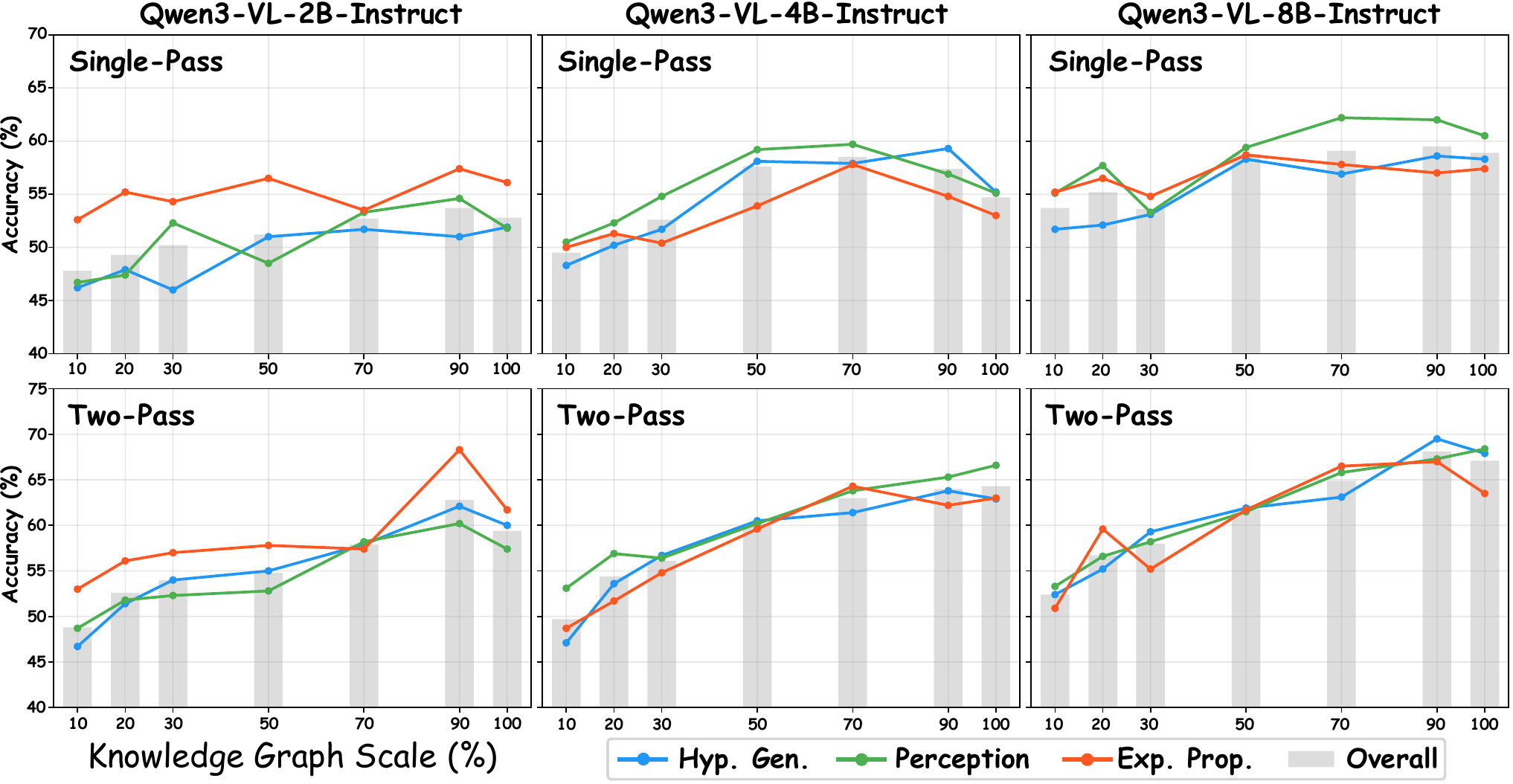}
\caption{Effect of MAPG scale on Qwen3-VL-Instruct (2B/4B/8B), Single-Pass and Two-Pass, on MicroVQA. Overall and task accuracy generally increase as the graph is randomly subsampled from 10\% to 100\%, with the largest gains between 10\% and 50\%. Some 90\% subgraphs outperform the full graph, potentially due to reduced retrieval noise.}
\label{fig:scale_study}
\end{figure}

We subsample the MAPG at six scales (10\%--90\%) and compare against the full graph; Figure~\ref{fig:scale_study} shows results across all three model sizes.
\textbf{(1)~Accuracy generally improves with scale.}
For all three model sizes, Two-Pass accuracy shows a clear upward trend from 10\% to 90\%, with Qwen3-VL-8B improving from 52.4\% (10\%) to 68.1\% (90\%) before settling at 67.1\% (100\%).
The largest gains occur between 10\% and 50\% (+9.3~pp for 8B), after which improvement flattens within this selected corpus.
\textbf{(2)~The 90\% scale sometimes surpasses 100\%.}
For 2B (62.8\% vs.\ 59.4\%) and 8B (68.1\% vs.\ 67.1\%), the 90\% graph slightly outperforms the full graph.
This may be attributed to the reduced noise from infrequent or low-quality entities in the remaining 10\%.
\textbf{(3)~Experiment Proposal benefits most from scale.}
This task shows the sharpest per-scale gains, while Perception grows more gradually, consistent with Experiment Proposal requiring broader contextual knowledge that denser graphs supply.

\subsection{Cross-benchmark evaluation: MicroBench}
\label{sec:microbench}

\begin{table*}[t]
\centering
\caption{
  Per-task accuracy (\%) on MicroBench.
  \textit{Coarse-grained} tasks: Modality, Sub-modality, Domain, Sub-domain, Stain.
  \textit{Fine-grained} tasks: Classification.
  Best result per model in column is \textbf{bolded}.
}
\label{tab:ubench}
\resizebox{\textwidth}{!}{%
\begin{tabular}{llccccc ccc}
\toprule
\multirow{2}{*}{\textbf{Model}} & \multirow{2}{*}{\textbf{Method}}
  & \multicolumn{5}{c}{\textit{Coarse-Grained Perception}}
  & \multicolumn{1}{c}{\textit{Fine-Grained}}
  & \multicolumn{2}{c}{\textit{Aggregate}} \\
\cmidrule(lr){3-7}\cmidrule(lr){8-8}\cmidrule(lr){9-10}
  & & Modality & Sub-mod. & Domain & Sub-dom. & Stain
    & Classif.
    & Coarse & Overall \\
\midrule
\multirow{3}{*}{Qwen3-VL-2B}
  & Baseline       & \textbf{79.5} & 47.2 & 51.4 & \textbf{57.7} & \textbf{49.1} & \textbf{34.3} & 57.3 & 53.8 \\
  & Single-Pass    & 63.4 & 46.2 & 53.9 & 53.2 & 46.4 & 32.5 & 52.8 & 49.7 \\
  & Two-Pass      & 77.5 & \textbf{59.2} & \textbf{56.3} & 54.1 & 45.8 & 33.8 & \textbf{59.0} & \textbf{55.2} \\
\midrule
\multirow{3}{*}{Qwen3-VL-4B}
  & Baseline       & 83.0 & 62.0 & 44.0 & \textbf{57.9} & 52.8 & 36.2 & 60.2 & 54.1 \\
  & Single-Pass    & \textbf{85.6} & \textbf{68.0} & \textbf{55.8} & 55.6 & \textbf{54.2} & 37.3 & \textbf{64.1} & \textbf{60.0} \\
  & Two-Pass      & 82.5 & 62.4 & 44.3 & \textbf{57.9} & 52.2 & \textbf{38.8} & 60.1 & 56.9 \\
\midrule
\multirow{3}{*}{Qwen3-VL-8B}
  & Baseline       & 80.7 & 67.1 & 53.7 & 52.6 & 56.9 & 34.3 & 62.4 & 58.1 \\
  & Single-Pass    & \textbf{83.8} & \textbf{69.8} & \textbf{67.1} & 52.6 & \textbf{61.3} & 33.8 & \textbf{67.2} & \textbf{62.1 }\\
  & Two-Pass      & 80.8 & 66.6 & 54.4 & \textbf{53.4} & 56.8 & \textbf{36.2} & 62.6 & 58.5 \\
\bottomrule
\end{tabular}%
}
\end{table*}

To assess transfer beyond MicroVQA, we evaluate on MicroBench~\citep{lozano2024micro}, which was not used in corpus filtering (Table~\ref{tab:ubench}).
\textsc{MicroWorld}'s gains concentrate in coarse-grained perception, where Single-Pass yields the strongest improvements at 4B (coarse: +3.9~pp, overall: 60.0\%, +5.9~pp) and 8B (+4.8~pp, 62.1\%, +4.0~pp), while Two-Pass leads at 2B (+1.7~pp, 55.2\%, +1.4~pp overall).
Fine-grained classification benefits are comparatively modest ($\leq$+2.6~pp at 4B), as this sub-task is more vision-centric and less amenable to external knowledge augmentation.
Each scale has a configuration that improves on baseline, but the gains depend on task and mode.

\newpage
\subsection{Corpus-selection controls}
\label{sec:corpus_controls}

\begin{table}[t]
\centering
\caption{Corpus-selection controls on MicroVQA.}
\label{tab:corpus_controls}
\begin{tabular}{lrrr}
\toprule
Graph & Def. coverage (\%) & Accuracy (\%) & Gain (pp) \\
\midrule
Random & 18.4 & 55.7 & +6.9 \\
Random (+ top-6K definitions) & 53.6 & 56.0 & +7.2 \\
Filtered (seed 42) & 82.8 & 59.3 & +10.5 \\
\bottomrule
\end{tabular}
\end{table}

Table~\ref{tab:corpus_controls} compares 5K-pair graphs (Qwen3-VL-8B, Two-Pass), with gains relative to the 48.8\% zero-shot baseline. Random sampling without MicroVQA contact retains +6.9--7.2~pp gains. Filtered (seed 42) reaches 59.3\%, but unequal definition coverage prevents attributing its advantage solely to corpus alignment (Appendix~\ref{app:corpus_controls}).
\subsection{Qualitative case study}
\label{sec:case_study}

We analyze Qwen3-VL-8B-Instruct on MicroVQA, examining how knowledge context (KC) flips incorrect answers to correct ones (\emph{upgrades}) and vice versa (\emph{downgrades}).
Full case illustrations are provided in Appendix~\ref{app:case_study}.

\paragraph{Upgrade mechanisms.}
We identify five recurring mechanisms through which KC improves reasoning (Figures~\ref{fig:upg1}--\ref{fig:upg5}):
\textbf{(1)~Terminology disambiguation}: KC provides precise definitions that help the model distinguish semantically similar options (e.g., ``dye specks'' vs.\ ``stain aggregates'' in Pap smear artifact identification);
\textbf{(2)~Domain-specific knowledge injection}: KC supplies entity-specific facts unavailable in the model's parametric memory (e.g., the role of sample purification in interpreting cryo-EM asymmetry of CHIKV);
\textbf{(3)~Reasoning level elevation}: KC elevates the model from local mechanistic explanations to broader biological insights (e.g., recognizing structural plasticity of alphavirus geometry rather than local protein conformational changes);
\textbf{(4)~Misconception correction}: KC corrects oversimplified textbook-level reasoning with domain-precise knowledge (e.g., lysosome ring arrangements reflecting multi-organellar defense assemblies rather than mere phagosome fusion);
\textbf{(5)~Logical chain correction}: KC rules out an incorrect causal chain (e.g., interpreting energy-intensive actin polymerization as energy conservation).

\paragraph{Downgrade mechanisms.}
We also identify five failure modes (Figures~\ref{fig:downg2}--\ref{fig:downg3}):
\textbf{(1)~Knowledge overriding visual evidence}: generic KC introduces textual noise that overrides the model's originally correct visual judgment;
\textbf{(2)~Context mismatch}: KC retrieved from a different experimental context (e.g., in vitro quality control advice) is misapplied to a distinct biological scenario (e.g., in vivo wound healing);
\textbf{(3)~Ambiguous KC signals}: conflicting information within the retrieved context leads the model to favor the more ``novel'' but incorrect explanation;
\textbf{(4)~Knowledge-induced over-reasoning}: detailed procedural knowledge causes the model to prefer a superficially direct but scientifically flawed strategy over a methodologically sound approach;
\textbf{(5)~Visual--textual confidence imbalance}: detailed textual knowledge leads the model to discount discriminative image features.
These failures motivate context relevance filtering, confidence-aware knowledge integration, and visual--textual evidence arbitration.

\section{Conclusion}

We presented \textsc{MicroWorld}, a framework that constructs a multimodal attributed property graph (MAPG) from scientific image--caption corpora and uses it to augment MLLM reasoning for microscopy VQA.
Graph-augmented prompting raises Qwen3-VL-8B to 67.1\% on MicroVQA, exceeding zero-shot GPT-5 (59.4\%); the identical context raises GPT-5 to 71.5\%.
The current gains are predominantly text-driven and strongest for knowledge-intensive reasoning. Cross-model transfer and smaller MicroBench and random-corpus gains support applicability beyond the main setting, with limits from corpus alignment, extraction noise, and fine-grained visual discrimination.
Qualitative analysis further reveals both the promise of knowledge-grounded reasoning and its failure modes (detailed in Appendix~\ref{app:case_study}), offering actionable insights for future systems. We also discuss ethics considerations in Appendix~\ref{supp:ethics}, discussions and more information in Appendix~\ref{supp:limi},~\ref{supp:data_leakage}.

\label{sec:main_end}
\clearpage
\begin{ack}

This work was supported by the National Key R\&D Program of China (Grant No. 2025YFF0514500) and Shanghai Municipal Special Program for Basic Research on General AI Foundation Models (Grant No. 2025SHZDZX025G11). The computations in this research were performed using the CFFF platform of Fudan University.

The authors declare no competing interests.

% This work was supported by the National Key R\&D Program of China (Grant No. 2025YFF0514500), NSFC (Grant No. 62372117 to W.T.), and Shanghai Municipal Special Program for Basic Research on General AI Foundation Models (Grant No. 2025SHZDZX025G11). The computations in this research were performed using the CFFF platform of Fudan University.

\end{ack}

\bibliographystyle{unsrt}  % plain 
\bibliography{references} 

\clearpage
\appendix

\section{Ethics}
\label{supp:ethics}

\paragraph{Ethical use of biomedical data.} Our work uses image--caption pairs sampled from OmniScience, which aggregates scientific figures and captions from biomedical literature. We do not introduce patient-level data or collect personally identifiable information. Access to a publication does not by itself establish permission to redistribute its figures.
\paragraph{Data licensing and usage.}
We intend to license eligible \textsc{MicroWorld}-derived assets under Creative Commons Attribution--NonCommercial--ShareAlike 4.0 (CC BY-NC-SA 4.0), subject to upstream permissions. This does not relicense source images or texts: their original attribution, copyright, and redistribution conditions remain applicable. Redistribution must preserve provenance and source-specific terms; where permission does not allow image redistribution, source identifiers and reconstruction instructions provide an alternative. Code licensing is separate from the data license. Users must verify the terms applicable to their intended use.
\section{Limitations, discussions and future work}
\label{supp:limi}

\paragraph{Text-centric knowledge representation.}
At the current 20K graph scale and corpus composition, \textsc{MicroWorld} is predominantly text-driven: the 8B joint text-only control loses only 1.1~pp on MicroVQA and 0.39~pp on MicroBench relative to full MAPG. Fine-grained morphological classification benefits less than terminology- and context-intensive reasoning (Section~\ref{sec:microbench}). Nearest-neighbor noise and the composite-figure nature of OmniScience limit the visual pathway; splitting figures into panels may improve specificity.

\paragraph{Benchmark-aligned corpus selection.}
MeSH-derived terms are weighted using MicroVQA question frequencies, so corpus selection is aligned with this benchmark even without accessing answers or options. The random-5K controls retain +6.9--7.2~pp gains, while filtered-5K reaches 59.3\% versus random-5K's 56.0\% and filtered-20K's 67.1\% (Appendix~\ref{app:corpus_controls}). The approximately 3.4~pp fixed-size gap also reflects unequal definition coverage (82.8\% versus 53.6\%). The approximately 7.8~pp filtered-5K-to-20K gap is likewise a descriptive scale comparison. Random-20K remains unevaluated. Transfer to distributions unlike MicroVQA is supported at smaller observed gains, including +4.0~pp on MicroBench at 8B Single-Pass.

\paragraph{Scale of \textsc{MicroWorld}.}
The graph uses 20K image--caption pairs from OmniScience because of computation and API-budget constraints. The scale study shows flattening only within this selected corpus. Expansion requires further extraction, indexing, storage, and embedding resources.

\paragraph{Limited Evaluation of Larger Open-Source MLLMs.}
Our open-source evaluation is restricted to Qwen3-VL and InternVL3.5 up to 8B parameters. GPT-5+MAPG improves further (71.5--73.3\%), but does not characterize larger open-source backbones or establish a universal interaction between model scale and external knowledge.

\paragraph{Retrieval quality bottleneck.}
Automatic relation extraction and definition generation can introduce errors, while entity disambiguation and image retrieval can add mismatched context. Our downgrade cases show that even factually valid knowledge may override correct visual judgments; current retrieval does not explicitly arbitrate conflicting evidence.

\paragraph{Broader impact.}
\textsc{MicroWorld} may help laboratories use external scientific knowledge without domain-specific fine-tuning. Its 8B result exceeds zero-shot GPT-5 on MicroVQA, while GPT-5 given the same context is stronger. With a prebuilt graph, open-source inference can run locally without sending query images to a commercial API; graph construction still uses GPT-5 for extraction and some definitions. Applying this pipeline to other scientific domains remains a direction for validation. The graph is automatically extracted from literature, so its outputs should support expert judgment rather than serve as authoritative scientific or clinical conclusions. Source-publication biases, extraction errors, and licensing restrictions also constrain responsible use.

\paragraph{Future work.}
Several promising directions emerge from the above limitations.
\textbf{First}, splitting composite figures into panels and indexing representative crops alongside textual descriptions may reduce visual retrieval noise.
\textbf{Second}, \emph{retrieval quality refinement} via reranking, retrieval-augmented filtering, or uncertainty-aware context selection would help mitigate the downgrade failure modes.
\textbf{Third}, scaling \textsc{MicroWorld} to the full OmniScience corpus with efficient approximate nearest-neighbor indexing is a natural next step to improve coverage.
Finally, evaluating larger open-source backbones, random-20K graphs, and controls with matched definition coverage would better characterize model-scale effects and corpus-alignment dependence.

\section{Concerns on data leakage}
\label{supp:data_leakage}

We distinguish benchmark distribution alignment from answer leakage, and clarify the provenance and nearest-neighbor semantics of image retrieval.

\paragraph{Corpus selection uses MicroVQA term frequencies.}

The weighting in Eq.~\ref{eq:query_weight} uses term frequencies from MicroVQA questions. This makes the selected corpus benchmark-aligned and limits claims about general microscopy transfer; it is separate from using ground-truth answers or answer options.

\textbf{(a) The vocabulary $\mathcal{T}$ originates from MeSH ontology, not from MicroVQA labels.}
The domain vocabulary $\mathcal{T}$ is derived entirely from biomedical entity mentions extracted via MeSH identifiers (a standardized biomedical ontology) and subsequently filtered by GPT-5 to retain microscopy-relevant terms (Appendix~\ref{app:corpus_selection}).
MicroVQA ground-truth answers and answer options are \emph{never} accessed during vocabulary construction.

\textbf{(b) Question frequencies create distribution alignment.}
$\mathrm{tf}_{\mathcal{Q}}(t)$ weights the domain vocabulary toward terms frequent in MicroVQA. MicroBench is not used in this filtering. The random-corpus controls characterize the dependence on corpus selection without eliminating the alignment confound (Table~\ref{tab:corpus_controls}; Appendix~\ref{app:corpus_controls}).

\paragraph{Nearest-neighbor image retrieval.}

The $\textsc{RetrieveSimilarImages}(x_q,\mathcal{G})$ step in Algorithm~\ref{alg:retrieval} encodes an unseen query and retrieves existing OmniScience image nodes by embedding similarity. It does not insert benchmark images, questions, answers, or options into the graph, nor require exact image identity.

\textbf{(a) MicroVQA images are in-house and independent of OmniScience.}
The MicroVQA benchmark~\citep{burgess2025microvqa} contains in-house laboratory images rather than images drawn from the OmniScience literature corpus.
OmniScience, by contrast, aggregates compositional figure panels extracted from open-access biomedical publications.
The MAPG image nodes are constructed from the selected OmniScience records, not from benchmark images. Their captions and linked entities supply external context for nearest-neighbor retrieval.

\paragraph{Summary.}
The graph is built from external literature without benchmark answers or options. Corpus selection nevertheless uses MicroVQA question frequencies and therefore aligns with its distribution.

\section{Method details}
\label{app:method_details}

\begin{algorithm}[t]
\caption{Visual--Semantic Subgraph Retrieval (Single-Pass / Two-Pass)}
\label{alg:retrieval}
\begin{algorithmic}[1]
\Require Query image $x_q$, question text $q$, graph $\mathcal{G}$, MLLM $\mathcal{F}$,
         parameters $M_{\text{text}}, M_{\text{vis}}, \rho_{\text{skip}}$,
         mode $\in \{\textsc{Single}, \textsc{Two}\}$
\Ensure Knowledge context string $\mathcal{C}$, answer $a$

\Statex \textbf{// --- Entity extraction (mode-dependent) ---}
\If{mode $=$ \textsc{Two}}
    \State $\mathcal{E}_q \gets \mathcal{F}.\textsc{ExtractEntities}(x_q, q)$
        \Comment{MLLM extracts up to 8 domain entities (Round~1)}
\Else
    \State $\mathcal{E}_q \gets \textsc{ScispaCy-NER}(q) \cup \textsc{NounChunks}(q)$
        \Comment{Offline NER for single-pass}
\EndIf

\Statex \textbf{// --- Textual pathway: match entities to KG ---}
\State $\mathcal{M}_{\text{text}} \gets \emptyset$
\For{$e \in \mathcal{E}_q$}
    \State $v \gets \textsc{AliasMatch}(e, \mathcal{G})$ \Comment{Exact alias lookup}
    \If{$v = \textsc{null}$}
        \State $v \gets \textsc{FuzzyMatch}(e, \mathcal{G},\, \rho_{\text{fuzzy}}{=}0.72)$
            \Comment{Fall back to Ratcliff/Obershelp fuzzy match}
    \EndIf
    \If{$v \neq \textsc{null} \,\land\, \text{freq}(v) \le \rho_{\text{skip}}$}
        \State $\mathcal{M}_{\text{text}} \gets \mathcal{M}_{\text{text}} \cup \{v\}$
    \EndIf
    \If{$|\mathcal{M}_{\text{text}}| \ge M_{\text{text}}$} \textbf{break} \EndIf
\EndFor

\Statex \textbf{// --- Visual pathway: nearest image nodes + graph expansion ---}
\State $\mathcal{I}_{\text{nn}} \gets \textsc{RetrieveSimilarImages}(x_q, \mathcal{G})$
    \Comment{Same encoder; nearest stored image vectors}
\State $\mathcal{I}_{\text{sim}} \gets \mathcal{I}_{\text{nn}} \cup \{v' \mid v \in \mathcal{I}_{\text{nn}},\, (v,\, \texttt{visually\_similar},\, v') \in \mathcal{E}_{\text{vsim}}\}$
\State $\mathcal{V}_{\text{vis}} \gets \emptyset$ \Comment{Candidate entity set from visual pathway}
\For{$v_{\text{img}} \in \mathcal{I}_{\text{sim}}$}
    \State $\mathcal{V}_{\text{vis}} \gets \mathcal{V}_{\text{vis}} \cup
           \{u \mid (v_{\text{img}},\, \texttt{shown\_in},\, u) \in \mathcal{E}_{\text{vis}}\}$
        \Comment{Collect caption entities via \texttt{shown\_in} edges}
\EndFor
\State $\mathcal{V}_{\text{vis}} \gets \textsc{RankByDescAvail}(\mathcal{V}_{\text{vis}})[{:}M_{\text{vis}}]$
    \Comment{$M_{\text{vis}}$ is the scalar cap; $\mathcal{V}_{\text{vis}}$ is the entity set}

\Statex \textbf{// --- Merge, expand, and format ---}
\State $\mathcal{M} \gets \mathcal{M}_{\text{text}} \cup \mathcal{V}_{\text{vis}}$
\For{$v \in \mathcal{M}$}
    \State Retrieve precomputed $\mathcal{N}_k(v)$ (Eq.~\ref{eq:adaptive_khop}) and descriptions $\boldsymbol{\psi}(\cdot)$
\EndFor
\State $\mathcal{C} \gets \textsc{FormatKnowledgeBlocks}(\mathcal{M}, \mathcal{N}_k, \boldsymbol{\psi})$

\Statex \textbf{// --- Answer generation ---}
\If{mode $=$ \textsc{Two}}
    \State $a \gets \mathcal{F}.\textsc{Answer}(x_q,\, \mathcal{C} \oplus q)$
        \Comment{Fresh single-turn prompt with retrieved context (Round~2)}
\Else
    \State $a \gets \mathcal{F}.\textsc{Answer}(x_q,\, \mathcal{C} \oplus q)$
\EndIf
\State \Return $\mathcal{C},\, a$
\end{algorithmic}
\end{algorithm}

\subsection{Entity type classification}
\label{app:type_classify}

A heuristic classifier assigns each entity to one of the five semantic types based on keyword matching and spaCy~\cite{honnibal2020spacy} entity labels:
\begin{equation}
  \tau(e) = \begin{cases}
    \textsc{Method}    & \text{if } e \cap \mathcal{K}_{\text{method}} \neq \emptyset, \\
    \textsc{Property}  & \text{if } e \cap \mathcal{K}_{\text{property}} \neq \emptyset, \\
    \textsc{Organism}  & \text{if label}(e) \in \{\textsc{Taxon, Species}\}, \\
    \textsc{Condition} & \text{if label}(e) \in \{\textsc{Disease, Phenotype}\}, \\
    \textsc{Structure} & \text{otherwise},
  \end{cases}
  \label{eq:type_classify}
\end{equation}
where $\mathcal{K}_{\text{method}}$ and $\mathcal{K}_{\text{property}}$ are curated keyword sets for experimental methods and morphological properties, respectively.
To ensure reproducibility, we provide part of the keyword sets below:
\begin{itemize}[leftmargin=2em, itemsep=2pt]
  \item $\mathcal{K}_{\text{method}}$: \emph{microscopy, imaging, staining, fluorescence, immunofluorescence, confocal, electron, cryo-EM, SEM, TEM, FISH, FRET, super-resolution, STORM, PALM, TIRF, phase-contrast, brightfield, widefield, two-photon, CRISPR, sequencing, PCR, Western blot, immunohistochemistry, flow cytometry, mass spectrometry, spectroscopy, tomography, diffraction, crystallography, NMR, patch-clamp, electrophysiology, transfection, transduction, knockout, knockdown, overexpression, fixation, permeabilization, embedding, sectioning}.
  \item $\mathcal{K}_{\text{property}}$: \emph{morphology, shape, size, length, width, diameter, thickness, density, intensity, brightness, contrast, texture, granularity, elongation, roundness, circularity, polarity, symmetry, distribution, localization, co-localization, overlap, pattern, structure, boundary, membrane, lumen, condensation, aggregation, fragmentation, branching, tubular, punctate, filamentous, ring-shaped, rod-shaped, spherical, irregular}.
\end{itemize}
Matching is performed case-insensitively on tokenized entity surface forms.
Entities matching neither $\mathcal{K}_{\text{method}}$ nor $\mathcal{K}_{\text{property}}$ keywords, and without a recognized Taxon/Species or Disease/Phenotype label, default to \textsc{Structure}.

\subsection{Vision--language embedding details}
\label{app:embedding_details}

We employ Qwen3-VL-Embedding~\citep{li2026qwen3} to compute $d$-dimensional embeddings for three modalities:

\begin{itemize}[leftmargin=2em, itemsep=2pt]
  \item \textbf{Image embeddings} $\mathbf{h}_{x} \in \mathbb{R}^d$: Each image $x_i$ is encoded with the instruction \emph{``Represent this microscopy image for knowledge graph node retrieval''}.
  \item \textbf{Mixed embeddings} $\mathbf{h}_{m} \in \mathbb{R}^d$: Each image--caption pair $(x_i, c_i)$ is jointly encoded with instruction \emph{``Represent this image-text pair for biomedical knowledge retrieval''}.
  \item \textbf{Entity embeddings} $\mathbf{h}_{e} \in \mathbb{R}^d$: Each entity canonical name is encoded with instruction \emph{``Represent this biomedical entity name for visual alignment''}.
\end{itemize}

All embeddings are $\ell_2$-normalized: $\boldsymbol{\phi}(v) = f_{\theta}(v) / \|f_{\theta}(v)\|_2$.

\subsection{K-hop neighbor precomputation}
\label{app:khop}

We precompute neighborhood structures using an adaptive $k$-hop expansion strategy based on node degree:
\begin{equation}
  \mathcal{N}_{k}(v) = \begin{cases}
    \mathcal{N}_1(v) \cup \mathcal{N}_2(v)  & \text{if } \deg(v) \le 3, \\
    \mathcal{N}_1(v)                         & \text{if } 3 < \deg(v) \le 20, \\
    \mathcal{N}_1(v)\big|_{\le 50}           & \text{if } \deg(v) > 20,
  \end{cases}
  \label{eq:adaptive_khop}
\end{equation}
where $\mathcal{N}_1(v)$ and $\mathcal{N}_2(v)$ denote the 1-hop and 2-hop neighbor sets of~$v$, and $|_{\le n}$ indicates truncation to at most~$n$ neighbors.
The three cases correspond to: \emph{(i)} sparse nodes that receive 2-hop expansion to ensure sufficient context; \emph{(ii)} medium-degree nodes ($4 \le \deg(v) \le 20$) whose full 1-hop neighborhood is used as-is; and \emph{(iii)} hub nodes ($\deg(v) > 20$) that are truncated to the 50 highest-scoring neighbors (ranked by the hybrid similarity score $S$, Eq.~\ref{eq:fused_score}) to prevent context overflow.

\subsection{Hybrid similarity ranking}
\label{app:hybrid_sim}

For each pair of entity nodes $(v_i, v_j)$, we compute the Jaccard~\cite{niwattanakul2013using} structural similarity over non-image 1-hop neighbors:
\begin{equation}
  J(v_i, v_j) = \frac{|\mathcal{N}(v_i) \cap \mathcal{N}(v_j)|}{|\mathcal{N}(v_i) \cup \mathcal{N}(v_j)|},
  \label{eq:jaccard}
\end{equation}
and the cosine embedding similarity $C(v_i, v_j) = \boldsymbol{\phi}(v_i)^{\top} \boldsymbol{\phi}(v_j)$.
The fused ranking score is:
\begin{equation}
  S(v_i, v_j) = \alpha \cdot J(v_i, v_j) + (1 - \alpha) \cdot C(v_i, v_j),
  \label{eq:fused_score}
\end{equation}
where $\alpha = 0.5$ by default. For each entity, we retain the top-$K$ most similar nodes.

\subsection{Graph augmentation with external corpora}
\label{app:graph_augment}

To scale knowledge coverage, we support merging the primary MAPG (\textsc{MicroWorld}) with a supplementary graph from the external corpus.
The merging follows three principles:
\textbf{(1)~primary-first}: nodes and descriptions from the primary graph are never overwritten;
\textbf{(2)~entity-only augmentation}: supplementary image nodes and \texttt{shown\_in} edges are excluded to preserve frequency statistics;
\textbf{(3)~embedding alignment}: embeddings from both graphs are concatenated with shared index files.
This yields $\mathcal{G}^{+}$ with broader entity coverage while maintaining visual retrieval integrity.

\subsection{Inference details}
\label{app:inference_details}

\paragraph{Entity frequency definition.}
\label{app:freq_def}
The entity frequency $\text{freq}(v)$ is the fraction of graph-construction image--caption pairs in which entity~$v$ appears:
\begin{equation}
  \text{freq}(v) = \frac{|\{d \in \mathcal{D} : v \in \text{Entities}(d)\}|}{|\mathcal{D}|},
  \label{eq:freq}
\end{equation}
where $\mathcal{D}$ is the set of all image-caption pairs used to construct the MAPG and $\text{Entities}(d)$ denotes the entity set extracted from image-caption pair~$d$.
Values are pre-computed over the 20k image-caption pairs corpus and stored in the entity registry.
The skip threshold $\rho_{\text{skip}}=0.08$ discards entities appearing in more than 8\% of corpus pairs (e.g.\ \emph{cells}, \emph{protein}); the compact threshold $\rho_{\text{compact}}=0.04$ triggers abbreviated representation for moderately common entities.

\paragraph{Textual pathway.}
We apply scispaCy NER to the question $q$, yielding candidate entity mentions.
Each mention is matched to the KG through:
\textbf{(1)}~exact alias lookup against the normalized alias index;
\textbf{(2)}~fuzzy matching via the Ratcliff/Obershelp algorithm~\cite{ratcliff1988gestalt} with cutoff~$\rho_{\text{fuzzy}}=0.72$.
Entities with corpus frequency exceeding $\rho_{\text{skip}}$ are discarded as overly generic.

\paragraph{Visual pathway.}
We encode $x_q$ with the same image encoder and instruction used for the stored 20K image nodes, then perform nearest-neighbor retrieval by cosine similarity. The returned node set is expanded over \texttt{visually\_similar} edges, and \texttt{shown\_in} links yield caption entities (Algorithm~\ref{alg:retrieval}). Entities are ranked by description availability and frequency, and at most $M_{\text{vis}}=3$ are added. An empty candidate set contributes no visual entities; the query itself need not be a stored node.

\paragraph{Context assembly.}
For each matched entity, we construct a knowledge block containing:
\textbf{(1)}~canonical name and semantic type;
\textbf{(2)}~NCBI/LLM-derived definition;
\textbf{(3)}~up to $M_{\text{nbr}}$ 1-hop neighbors with relation types and descriptions;
\textbf{(4)}~2-hop context for low-degree nodes ($|\mathcal{N}_1(v)| \le 3$), capped by \texttt{max\_2hop}.

\paragraph{Frequency-aware formatting.}
Entities with moderate frequency ($\rho_{\text{compact}} < \text{freq}(v) \le \rho_{\text{skip}}$) receive compact representations (truncated to 120~characters).

\begin{table}[t]
\centering
\caption{Complete relation schema of the multimodal attributed property graph.}
\label{tab:relations}

\begin{tabular}{llll}
\toprule
\textbf{Relation Type} & \textbf{Source} & \textbf{Direction} & \textbf{Semantics} \\
\midrule
\texttt{observed\_by}      & LLM extraction & Directed   & Structure $\to$ imaging method \\
\texttt{has\_property}     & LLM extraction & Directed   & Entity $\to$ morphological property \\
\texttt{located\_in}       & LLM extraction & Directed   & Entity $\to$ spatial location \\
\texttt{interacts\_with}   & LLM extraction & Directed   & Functional/physical interaction \\
\texttt{part\_of}          & LLM extraction & Directed   & Sub-component $\to$ whole \\
\texttt{co\_occurs\_with}  & Co-occurrence  & Undirected & Same-caption co-occurrence \\
\texttt{shown\_in}         & Visual link    & Directed   & Image $\to$ depicted entity \\
\texttt{visually\_similar} & Embedding sim. & Undirected & Image visual similarity ($\ge\sigma$) \\
\bottomrule
\end{tabular}
\end{table}

\section{RAG baseline results at 2B and 4B scales}
\label{app:rag_2b4b}

Table~\ref{tab:rag_comparison_2b4b} reports the full RAG baseline comparisons at 2B and 4B model scales, complementing the 8B results in Table~\ref{tab:rag_comparison} of the main text.

\begin{table}[ht]
\centering
\caption{Comparison with RAG baselines at 2B and 4B scales (best top-$k$ per method). Type: T=text, V=vision, T+V=text-vision, G=graph. \textsc{MicroWorld} uses Two-Pass mode. Best results per column within each scale are \textbf{bolded}.}
\label{tab:rag_comparison_2b4b}

\setlength{\tabcolsep}{4pt}
\begin{tabular}{llccccc}
\toprule
\textbf{Method} & \textbf{Type} & \textbf{$k$} & \textbf{Hyp.~Gen.} & \textbf{Perception} & \textbf{Exp.~Prop.} & \textbf{Overall} \\
\midrule
\multicolumn{7}{l}{\textit{2B scale}} \\
\rowcolor{rowgray}
\textsc{MicroWorld} (Ours) & G+V+T & --  & \textbf{60.0} & \textbf{57.4} & \textbf{61.7} & \textbf{59.4} \\
FLMR              & T     & 3   & 41.9          & 43.6          & 50.0          & 44.3          \\
VisRAG            & V     & 3   & 45.6          & 42.2          & 50.2          & 45.3          \\
ColPali           & V     & 5   & 41.0          & 45.3          & 60.4          & 46.9          \\
REVEAL            & T+V   & 1   & 45.5          & 44.6          & 50.0          & 46.2          \\
GraphRAG          & G     & 1   & 44.0          & 46.7          & 47.8          & 45.9          \\
LightRAG          & G     & 3   & 40.0          & 42.6          & 54.3          & 44.1          \\
\midrule
\multicolumn{7}{l}{\textit{4B scale}} \\
\rowcolor{rowgray}
\textsc{MicroWorld} (Ours) & G+V+T & --  & \textbf{62.9} & \textbf{66.6} & \textbf{63.0} & \textbf{64.3} \\
FLMR              & T     & 5   & 44.2          & 47.7          & 46.5          & 46.0          \\
VisRAG            & V     & 5   & 41.7          & 46.2          & 49.5          & 45.1          \\
ColPali           & V     & 3   & 44.8          & 48.2          & 48.3          & 46.8          \\
REVEAL            & T+V   & 3   & 43.8          & 48.0          & 47.8          & 46.3          \\
GraphRAG          & G     & 3   & 46.7          & 48.5          & 48.3          & 47.7          \\
LightRAG          & G     & 5   & 45.0          & 51.5          & 45.7          & 47.6          \\
\bottomrule
\end{tabular}
\end{table}

\section{Inference latency analysis}
\label{app:latency}

\textsc{MicroWorld}'s Two-Pass inference pipeline introduces an entity extraction step before the main knowledge-augmented reasoning call.
We benchmark its end-to-end per-query latency against a Single-Pass baseline to characterize the practical overhead and the effect of entity extraction batch size on latency and throughput.

\paragraph{Setup.}
All experiments are conducted on a single NVIDIA RTX 3090 24\,GB GPU.
\emph{Single-Pass} runs spaCy before standard MLLM inference.
\emph{Two-Pass} first performs a batched entity extraction call (on average 6--8 biomedical entities per query), retrieves the relevant KG subgraph, and then performs the knowledge-augmented reasoning call.
We sweep the entity extraction batch size $B \in \{1, 2, 4, 8, 16\}$ and report mean latency, 95th-percentile (P95) latency, and throughput in queries per second (QPS).

\paragraph{Results.}
Table~\ref{tab:latency} reports the full latency sweep.
Three findings stand out.
\textbf{(1) Batching drastically reduces Two-Pass overhead.}
At $B{=}1$, Two-Pass processes entities sequentially and incurs a $2.9$--$3.9\times$ mean-latency penalty over Single Pass.
Increasing $B$ to 4 brings mean latency within ${\sim}25\%$ of Single Pass for the 2B and 4B models; the 8B overhead is 38.6\%.
\textbf{(2) Two-Pass surpasses Single-Pass throughput at large batch sizes.}
At $B{=}8$, Two-Pass reaches 2.11\,QPS (2B) and 2.04\,QPS (4B), outperforming the corresponding Single-Pass throughput (1.54 and 1.52\,QPS) by ${\sim}34\%$.
This counterintuitive result arises because a single batched entity-extraction call amortizes prompt-processing overhead over multiple entities, making the two-LLM-call pipeline cheaper per query than a single un-batched call when the batch is large enough.
\textbf{(3) Model size has limited impact on Single-Pass latency.}
Single-Pass mean latency is nearly identical across the 2B, 4B, and 8B variants (649--660\,ms), suggesting that, under this hardware configuration, the bottleneck lies in memory bandwidth rather than arithmetic throughput for these model sizes.

\begin{table}[t]
\centering
\footnotesize	
\caption{%
  \textbf{Inference latency comparison between Single-Pass and Two-Pass on MicroVQA ($B$ = entity extraction batch size).}
  Mean and P95 latency (ms) and throughput (QPS) are reported over 64 test samples on a single NVIDIA RTX 3090 24\,GB GPU.
  ``---'' indicates GPU out of memory (OOM).
  Two-Pass at $B{=}8$ already matches or exceeds Single-Pass throughput.
}
\label{tab:latency}
\setlength{\tabcolsep}{5pt}
\begin{tabular}{llccccccccc}
\toprule
\multirow{2}{*}{Method} & \multirow{2}{*}{$B$} &
  \multicolumn{3}{c}{Qwen3-VL-2B} &
  \multicolumn{3}{c}{Qwen3-VL-4B} &
  \multicolumn{3}{c}{Qwen3-VL-8B} \\
\cmidrule(lr){3-5}\cmidrule(lr){6-8}\cmidrule(lr){9-11}
& & Mean\,(ms) & P95\,(ms) & QPS
  & Mean\,(ms) & P95\,(ms) & QPS
  & Mean\,(ms) & P95\,(ms) & QPS \\
\midrule
\rowcolor{rowgray}
Single Pass & --- & 649.6 & 1339.3 & 1.54 & 659.6 & 1338.4 & 1.52 & 651.3 & 1338.4 & 1.54 \\
\midrule
Two-Pass & 1  & 2503.9 & 2916.8 & 0.40 & 1930.2 & 2568.6 & 0.52 & 2227.7 & 2818.8 & 0.45 \\
Two-Pass & 2  & 1280.2 & 1652.2 & 0.78 & 1332.3 & 1664.0 & 0.75 & 1526.8 & 1828.0 & 0.65 \\
Two-Pass & 4  &  789.1 & 1618.0 & 1.27 &  790.4 & 1475.7 & 1.27 &  902.6 & 1274.2 & 1.11 \\
Two-Pass & 8  &  473.3 &  834.8 & \textbf{2.11} &  489.9 &  793.8 & \textbf{2.04} & --- & --- & --- \\
Two-Pass & 16 &  278.3 &  340.8 & \textbf{3.59} & --- & --- & --- & --- & --- & --- \\
\bottomrule
\end{tabular}
\end{table}

\section{Domain-relevant corpus selection from OmniScience}
\label{app:corpus_selection}

The OmniScience corpus contains a broad collection of biological science image--caption pairs spanning diverse sub-fields.
To construct a supplementary knowledge graph that is maximally relevant to microscopy reasoning, we design a weighted TF-IDF~\cite{sparck1972statistical} scoring algorithm to select a compact, domain-focused subset from OmniScience.
The core idea is to prioritize image-caption pairs that densely cover the specialized vocabulary of the microscopy domain (as reflected in MeSH identifiers aided with GPT-5) while penalizing image-caption pairs dominated by generic biomedical terms.

\paragraph{Terminology extraction.}
We first extract a domain vocabulary $\mathcal{T} = \{t_1, t_2, \ldots, t_{|\mathcal{T}|}\}$ by collecting all biomedical entity mentions from the MeSH identifiers filtered by GPT-5.
Representative high-frequency terms include \emph{cells}, \emph{microscopy}, \emph{protein}, \emph{fluorescence}, \emph{staining}, \emph{mitochondrial}, and \emph{confocal}, which constitute the core vocabulary of microscopic sciences.

\paragraph{Query-side term weighting.}
For each term $t \in \mathcal{T}$, we compute a query-side weight that balances its importance in the target domain against its generality in the source corpus:
\begin{equation}
  w_t = \mathrm{tf}_{\mathcal{Q}}(t) \times \mathrm{IDF}_{\mathcal{S}}(t), \quad \text{where} \quad \mathrm{IDF}_{\mathcal{S}}(t) = \log \frac{|\mathcal{S}|}{1 + |\{d \in \mathcal{S} : t \in d\}|},
  \label{eq:query_weight}
\end{equation}
where $\mathrm{tf}_{\mathcal{Q}}(t)$ is the term frequency of $t$ in the MicroVQA question set $\mathcal{Q}$, $\mathcal{S}$ denotes the full OmniScience corpus, and $|\mathcal{S}|$ is the total number of image-caption pairs.
This formulation assigns higher weights to terms that are both frequent in MicroVQA and rare in OmniScience, thereby emphasizing domain-specific terminology over generic biomedical vocabulary.

\begin{figure}[t]
\centering
\includegraphics[width=1\textwidth]{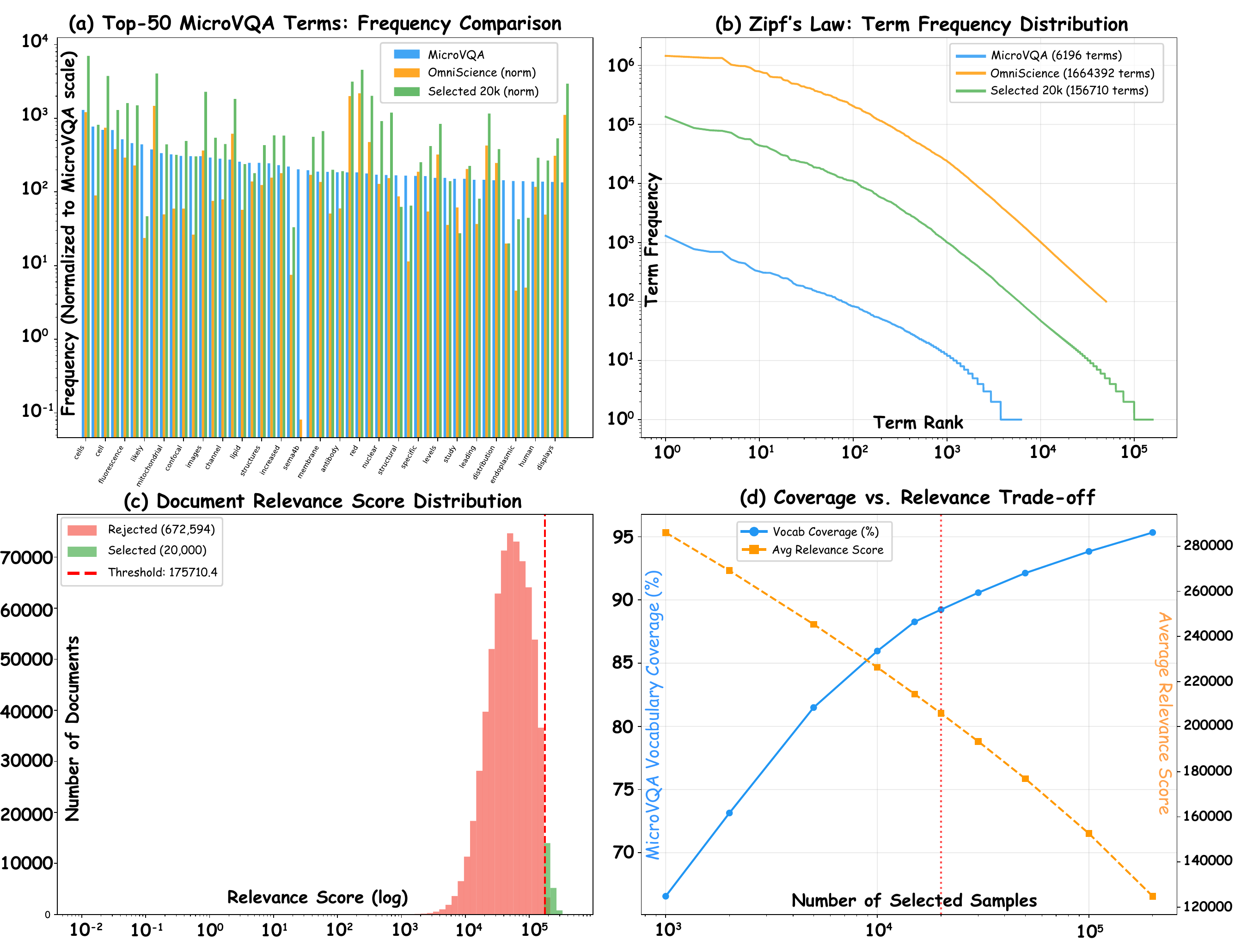}
\caption{\textbf{Corpus filtering analysis for \textsc{MicroWorld} subset selection.}
\textbf{(a) Top-50 MicroVQA Terms: Frequency Comparison.}
Term frequency distributions of the top-50 MicroVQA terms across three corpora, MicroVQA questions, the full OmniScience corpus, and the filtered 20k subset, showing that the selected subset closely mirrors the MicroVQA term distribution whereas the unfiltered corpus diverges substantially.
\textbf{(b) Zipf's Law: Term Frequency Distribution.}
Rank--frequency plots for the three corpora, confirming that the filtered subset preserves the power-law rank--frequency profile of MicroVQA and does not distort its linguistic statistics.
\textbf{(c) Image-caption pairs Relevance Score Distribution.}
Histogram of TF-IDF-based relevance scores over all OmniScience image-caption pairs, with the selection threshold (top-$K=20{,}000$) indicated; the heavy-tailed distribution motivates threshold-based selection to retain only domain-aligned image-caption pairs.
\textbf{(d) Coverage vs.\ Relevance Trade-off.}
MicroVQA vocabulary coverage as a function of subset size $K$; $K=20{,}000$ achieves 89.2\% coverage of MicroVQA domain vocabulary while maintaining corpus compactness, representing a favorable balance between knowledge breadth and indexing cost.}
\label{fig:omniscience_filter}
\end{figure}

\paragraph{Image-caption Pairs scoring.}
For each candidate image-caption pair $d \in \mathcal{S}$, we compute a relevance score by aggregating the weighted contributions of all matching domain terms, modulated by a diversity bonus:
\begin{equation}
  \mathrm{score}(d) = \underbrace{\sum_{t \,\in\, d \,\cap\, \mathcal{T}} w_t \cdot \log\bigl(1 + \mathrm{tf}_d(t)\bigr)}_{\text{weighted term relevance}} \;\times\; \underbrace{\log\bigl(|\{t \in \mathcal{T} : t \in d\}| + 1\bigr)}_{\text{diversity bonus}},
  \label{eq:doc_score}
\end{equation}
where $\mathrm{tf}_d(t)$ is the local term frequency of $t$ in image-caption pair $d$.
The logarithmic dampening of local term frequency prevents individual high-frequency terms from dominating the score.
The diversity bonus $\log(|\{t \in \mathcal{T} : t \in d\}| + 1)$ encourages the selection of image-caption pairs that cover a broader range of MicroVQA concepts, favoring comprehensiveness over narrow specialization.

\paragraph{Subset selection.}
We rank all OmniScience image-caption pairs by their relevance score in descending order and select the top-$K$ image-caption pairs to form the filtered subset $\mathcal{S}^{*} \subset \mathcal{S}$:
\begin{equation}
  \mathcal{S}^{*} = \operatorname*{top\text{-}K}_{d \,\in\, \mathcal{S}} \;\mathrm{score}(d).
  \label{eq:subset_select}
\end{equation}
In practice, we set $K = 20{,}000$.
Figure~\ref{fig:omniscience_filter} provides a detailed analysis of the filtering outcome across four dimensions:
\textbf{(1)}~the top-50 MicroVQA term frequency comparison across the three corpora (MicroVQA, full OmniScience, and the selected 20k subset), demonstrating that the filtered subset exhibits a term distribution substantially closer to MicroVQA than the unfiltered corpus;
\textbf{(2)}~the Zipf's law distribution~\cite{zipf2013psycho,zipf2016human}, confirming that the filtered corpus preserves the rank--frequency profile of MicroVQA;
\textbf{(3)}~the relevance score distribution with the selection threshold;
and \textbf{(4)}~the vocabulary coverage--relevance trade-off curve, showing that $K = 20{,}000$ achieves 89.2\% coverage of the MicroVQA domain vocabulary, representing a favorable balance between knowledge breadth and corpus compactness.

\section{Case study}
\label{app:case_study}

To gain deeper insight into how \textsc{MicroWorld}'s knowledge context (KC) influences model reasoning beyond aggregate accuracy numbers, we conduct a systematic qualitative analysis of Qwen3-VL-8B-Instruct on MicroVQA under the Two-Pass inference paradigm.
Specifically, we compare the model's predictions with and without KC augmentation, and categorize every answer flip into one of two groups:
\emph{upgrades} (baseline incorrect $\to$ KC-augmented correct) and \emph{downgrades} (baseline correct $\to$ KC-augmented incorrect).
From the full set of flipped answers, we select five representative cases from each group that illustrate distinct and recurring mechanisms.

For each case, we present the original question, the microscopy image, the baseline and KC-augmented predictions, the relevant knowledge retrieved from the MAPG, and a mechanistic analysis explaining \emph{why} the knowledge helped or hurt.
The upgrade cases illustrate five mechanisms: \emph{terminology disambiguation}, \emph{domain-specific knowledge injection}, \emph{reasoning level elevation}, \emph{misconception correction}, and \emph{logical chain correction}.
The downgrade cases illustrate five failure modes: \emph{knowledge overriding visual evidence}, \emph{context mismatch}, \emph{ambiguous KC signals}, \emph{knowledge-induced over-reasoning}, and \emph{visual--textual confidence imbalance}.
These patterns provide actionable insights for improving knowledge-augmented reasoning systems in future work.

\subsection{Upgrade cases: Knowledge enabling correct reasoning}

\begin{figure}[H]
  \centering
  \includegraphics[width=1\textwidth]{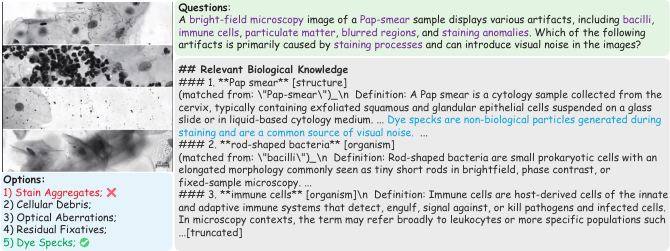}
  \caption{\textbf{Upgrade Case~1} (Perception): \emph{Terminology disambiguation.}
  \textbf{Question}: In a Pap-smear bright-field image, which artifact is primarily caused by staining and introduces visual noise?
  \textbf{Baseline} selects Option~1 (Stain Aggregates) \ding{55};
  \textbf{KC-augmented} selects Option~5 (Dye Specks) \ding{51}.
  The KG provides: \emph{``Dye specks are non-biological particles generated during the staining process and are a common source of visual noise.''}
  While the baseline hesitates between two semantically similar options and defaults to the more visually prominent one, KC supplies a precise terminological definition that resolves the ambiguity.}
  \label{fig:upg1}
\end{figure}

\begin{figure}[H]
  \centering
  \includegraphics[width=1\textwidth]{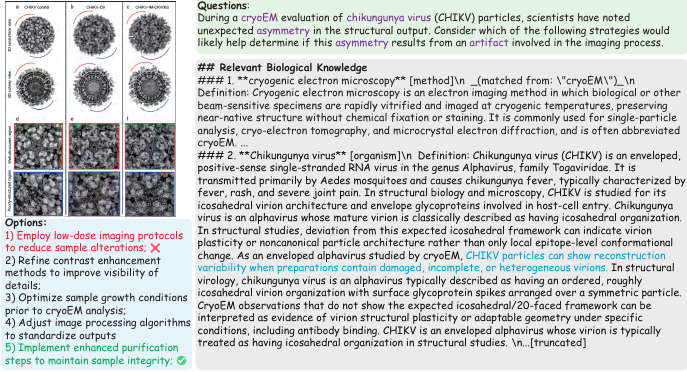}
  \caption{\textbf{Upgrade Case~2} (Experiment Proposal): \emph{Domain-specific knowledge injection.}
  \textbf{Question}: Unexpected asymmetry observed in cryo-EM of Chikungunya virus (CHIKV), which strategy best determines whether this is an artifact?
  \textbf{Baseline} selects Option~1 (low-dose imaging to reduce radiation damage) \ding{55};
  \textbf{KC-augmented} selects Option~5 (enhanced purification to maintain sample integrity) \ding{51}.
  The KG provides CHIKV-specific knowledge: \emph{``CHIKV particles can show reconstruction variability when preparations contain damaged, incomplete, or heterogeneous virions. Sample quality and purification are therefore important controls.''}
  The baseline applies generic cryo-EM reasoning (radiation damage as the default artifact source), whereas KC injects entity-specific facts indicating that sample purification quality is the key control variable for this enveloped virus.}
  \label{fig:upg2}
\end{figure}

\begin{figure}[H]
  \centering
  \includegraphics[width=1\textwidth]{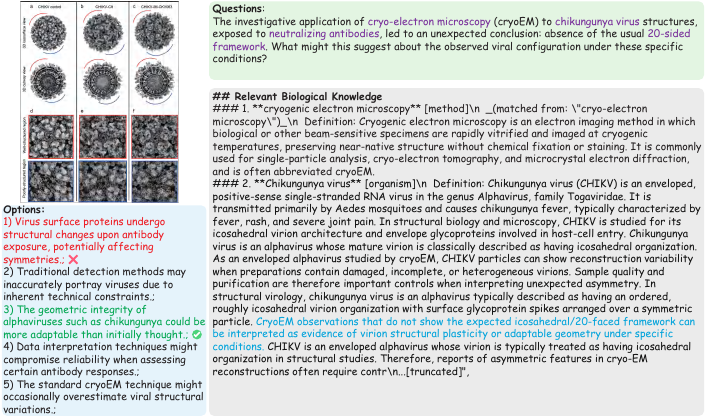}
  \caption{\textbf{Upgrade Case~3} (Perception): \emph{Reasoning level elevation.}
  \textbf{Question}: After neutralizing antibody treatment, CHIKV loses icosahedral symmetry. What does this imply?
  \textbf{Baseline} selects Option~1 (surface protein conformational changes affect symmetry) \ding{55};
  \textbf{KC-augmented} selects Option~3 (alphavirus geometric integrity is more adaptive than previously thought) \ding{51}.
  The baseline produces a locally correct but suboptimal ``mechanistic'' explanation (protein conformational change), while KC's alphavirus structural knowledge elevates reasoning to a higher-level biological insight, that viral geometric integrity itself exhibits structural plasticity, rather than being merely a consequence of local protein changes.}
  \label{fig:upg3}
\end{figure}

\begin{figure}[H]
  \centering
  \includegraphics[width=1\textwidth]{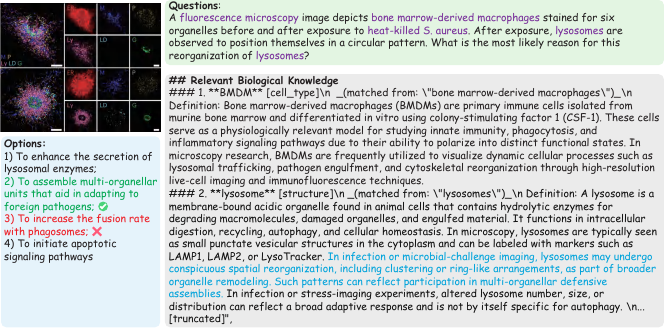}
  \caption{\textbf{Upgrade Case~4} (Hypothesis Generation): \emph{Misconception correction.}
  \textbf{Question}: After macrophage exposure to heat-killed \emph{S.~aureus}, lysosomes exhibit ring-like arrangements. What is the underlying cause?
  \textbf{Baseline} selects Option~3 (increased phagosome fusion rate) \ding{55};
  \textbf{KC-augmented} selects Option~2 (assembly of multi-organellar cooperative defense units) \ding{51}.
  The KG definition of lysosomes explicitly states: \emph{``ring-like arrangements reflect participation in multi-organellar defensive assemblies and should not automatically be interpreted only as increased lysosome--phagosome fusion.''}
  The baseline relies on textbook-level reasoning (lysosome reorganization $\approx$ increased fusion), while KC provides a domain-precise correction that the ring pattern reflects broader organelle remodeling for cooperative defense.}
  \label{fig:upg4}
\end{figure}

\begin{figure}[H]
  \centering
  \includegraphics[width=1\textwidth]{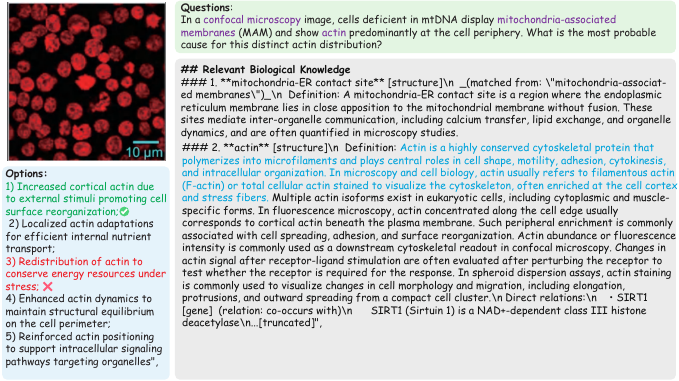}
  \caption{\textbf{Upgrade Case~5} (Hypothesis Generation): \emph{Logical chain correction.}
  \textbf{Question}: In mtDNA-deficient cells, actin concentrates at the cell periphery. What is the most likely cause?
  \textbf{Baseline} selects Option~3 (actin redistribution to conserve energy) \ding{55};
  \textbf{KC-augmented} selects Option~1 (external stimuli promote cell surface remodeling, increasing cortical actin) \ding{51}.
  The baseline follows the plausible but incorrect causal chain: mitochondrial dysfunction $\to$ energy deficit $\to$ energy conservation.
  KC about actin and cytoskeleton reveals that actin polymerization is itself energy-intensive, invalidating the ``conservation'' hypothesis.
  The correct explanation is that cortical actin enrichment reflects normal cell surface remodeling, which KC helps the model recognize.}
  \label{fig:upg5}
\end{figure}

\subsection{Downgrade cases: Knowledge inducing incorrect reasoning}

While \textsc{MicroWorld} improves overall accuracy substantially, a non-trivial fraction of answer flips go in the wrong direction.
Understanding these failure modes is crucial for guiding future improvements.
The five cases below illustrate how retrieved knowledge can mislead reasoning through different mechanisms, from overriding correct visual judgments with irrelevant textual noise to misapplying knowledge from a different experimental context.
Notably, several of these failures stem not from incorrect knowledge in the graph, but from the model's inability to properly weigh, contextualize, or arbitrate between visual evidence and textual knowledge.

\begin{figure}[H]
  \centering
  \includegraphics[width=1\textwidth]{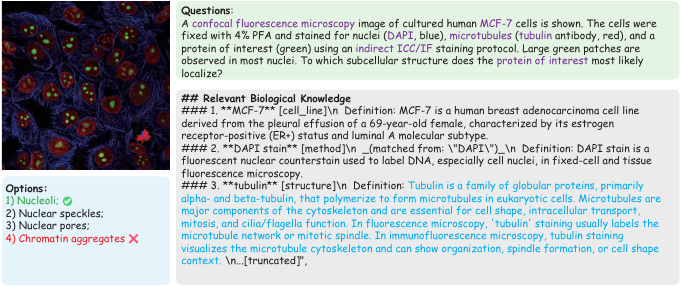}
  \caption{\textbf{Downgrade Case~1} (Perception): \emph{Knowledge overriding visual evidence.}
  \textbf{Question}: Identify the subcellular localization of green patches within MCF-7 cell nuclei.
  \textbf{Baseline} correctly selects Option~1 (Nucleoli) \ding{51}, based on direct visual assessment of bright, discrete intranuclear foci consistent with nucleolar morphology.
  \textbf{KC-augmented} incorrectly selects Option~4 (Chromatin aggregates) \ding{55}.
  KC contains no information about the specific protein's localization, but provides generic knowledge about MCF-7 cells, DAPI staining, and confocal microscopy.
  This textual noise causes the model to second-guess its visual judgment, reinterpreting the signal pattern as ``not confined to discrete structures'' and switching to chromatin aggregates.
  \textbf{Lesson}: when KC lacks query-relevant knowledge, generic context can erode the model's visual confidence.}
  \label{fig:downg2}
\end{figure}

\begin{figure}[H]
  \centering
  \includegraphics[width=1\textwidth]{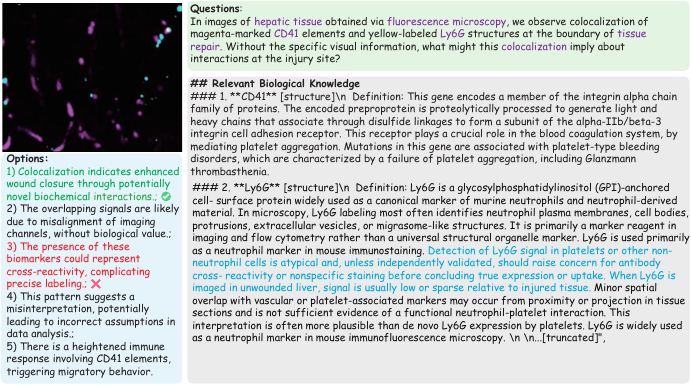}
  \caption{\textbf{Downgrade Case~2} (Perception): \emph{Context mismatch.}
  \textbf{Question}: CD41/Ly6G co-localization at a hepatic injury border, what does it indicate?
  \textbf{Baseline} correctly selects Option~1 (novel biochemical interaction enhancing wound healing) \ding{51}.
  \textbf{KC-augmented} incorrectly selects Option~3 (antibody cross-reactivity artifacts) \ding{55}.
  KC states: \emph{``Detection of Ly6G signal in platelets should raise concern for antibody cross-reactivity.''}
  This advice originates from in vitro platelet preparation quality control, but the question concerns in vivo tissue repair imaging, where CD41--Ly6G co-localization reflects genuine platelet--neutrophil cooperative repair.
  The model misapplies quality-control guidance from a different experimental context.}
  \label{fig:downg4}
\end{figure}

\begin{figure}[H]
  \centering
  \includegraphics[width=1\textwidth]{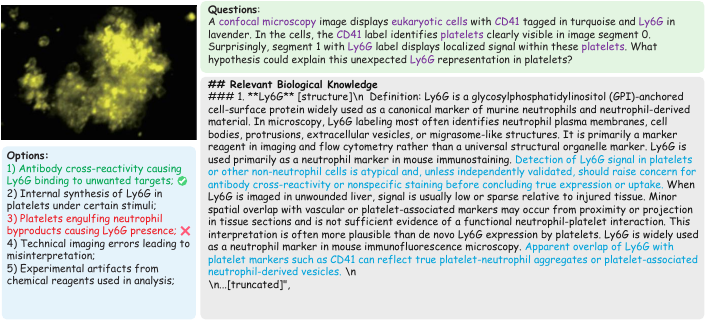}
  \caption{\textbf{Downgrade Case~3} (Hypothesis Generation): \emph{Ambiguous KC signals.}
  \textbf{Question}: Unexpected Ly6G signal detected in isolated platelets, what is the most likely explanation?
  \textbf{Baseline} correctly selects Option~1 (antibody cross-reactivity) \ding{51}.
  \textbf{KC-augmented} incorrectly selects Option~3 (platelet phagocytosis of neutrophil-derived material) \ding{55}.
  KC simultaneously contains: (i)~\emph{``should raise concern for antibody cross-reactivity''} (supporting Option~1) and (ii)~\emph{``can reflect true platelet--neutrophil aggregates or platelet-associated neutrophil-derived vesicles''} (supporting Option~3).
  Faced with conflicting signals, the model favors the more ``novel'' biological explanation from the KG over the more fundamental but correct answer.}
  \label{fig:downg5}
\end{figure}

\begin{figure}[H]
  \centering
  \includegraphics[width=1\textwidth]{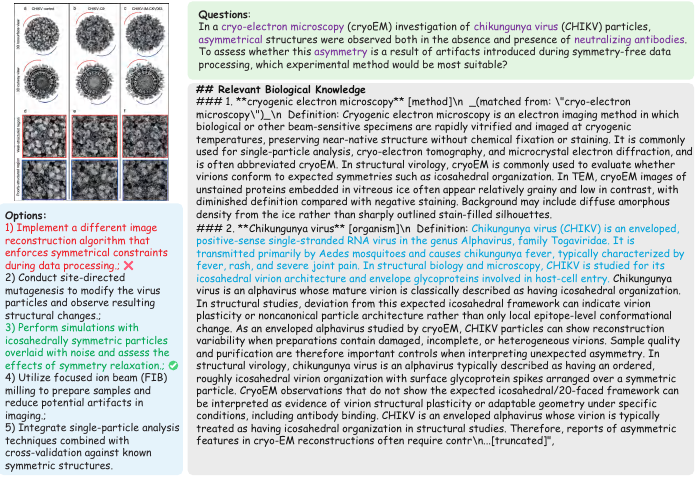}
  \caption{\textbf{Downgrade Case~4} (Experiment Proposal): \emph{Knowledge-induced over-reasoning.}
  \textbf{Question}: How to verify whether the observed cryo-EM asymmetry in CHIKV is an artifact?
  \textbf{Baseline} correctly selects Option~3 (simulate icosahedral particles with added noise to evaluate symmetry relaxation effects) \ding{51}.
  \textbf{KC-augmented} incorrectly selects Option~1 (use reconstruction algorithms with enforced symmetry constraints) \ding{55}.
  The baseline correctly reasons that forward simulation is the methodologically sound approach to test the artifact hypothesis.
  KC provides detailed cryo-EM processing knowledge, leading the model to prefer a superficially more direct strategy (enforced symmetry).
  However, enforcing symmetry only eliminates asymmetry without determining its origin, a critical distinction between ``testing a hypothesis'' and ``avoiding the problem.''}
  \label{fig:downg1}
\end{figure}

\begin{figure}[H]
  \centering
  \includegraphics[width=1\textwidth]{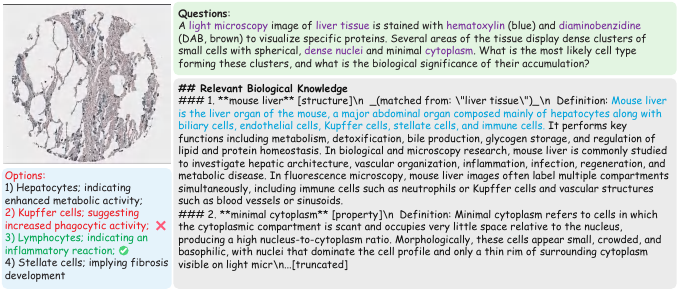}
  \caption{\textbf{Downgrade Case~5} (Perception): \emph{Visual--textual confidence imbalance.}
  \textbf{Question}: Identify small, densely clustered cells in liver tissue.
  \textbf{Baseline} correctly selects Option~3 (lymphocytes indicating inflammatory response) \ding{51}, accurately recognizing morphological features: small, round, high nuclear-to-cytoplasmic ratio, and dense clustering.
  \textbf{KC-augmented} incorrectly selects Option~2 (Kupffer cells indicating increased phagocytic activity) \ding{55}.
  KC provides detailed knowledge about Kupffer cells (liver-resident macrophages), which induces the model to reinterpret the observed cells as Kupffer cells despite their morphology being inconsistent (Kupffer cells are larger and irregularly shaped).
  The presence of domain knowledge reduces the model's trust in its own visual evidence.}
  \label{fig:downg3}
\end{figure}

\clearpage

\section{Supplementary experimental controls}
\label{app:rebuttal_evidence}

This appendix details the corpus-selection protocol and completed controls for graph retrieval, backbone transfer, and knowledge presentation.

\subsection{Corpus-selection protocol}
\label{app:corpus_controls}

Table~\ref{tab:corpus_controls} compares 5K-pair graphs on MicroVQA using Qwen3-VL-8B, Two-Pass. We uniformly sampled 5K image--caption pairs from OmniScience without MicroVQA contact and reran triplet extraction, graph construction, definition generation, and retrieval-augmented inference. Random uses the initial definition set; Random (+ top-6K definitions) adds definitions for the top-6K corpus-frequent entities in the same graph. Filtered (seed 42) uniformly subsamples the vocabulary-filtered 20K corpus, rebuilding the graph from the sampled pairs' original GPT-5 triplets and reusing their definitions. Both random conditions use no MicroVQA information; the filtered condition inherits benchmark-aligned corpus selection.

All gains use the 48.8\% zero-shot reference. The unrounded filtered and definition-enriched random accuracies are 59.31\% and 55.95\%, respectively, a 3.36-point difference. Their unequal definition coverage (82.8\% versus 53.6\%) prevents attributing this difference solely to vocabulary alignment. The approximately 7.8-point gap between filtered-5K and the full filtered-20K result (67.1\%) is a descriptive scale comparison. These controls support benefits beyond benchmark-aligned selection but do not isolate its causal contribution.

\subsection{2-hop and visual-edge control}
\label{app:factorial}

We vary the 2-hop neighbor budget (\texttt{max\_2hop}) and \texttt{visually\_similar} edges in a complete $2\times2$ control on MicroVQA (1,042 questions, Qwen3-VL-8B, Two-Pass). A budget of 3 caps the number of 2-hop neighbors; it does not denote three-hop traversal.

\begin{table}[ht]
\centering
\caption{2-hop $\times$ visual-edge control.}
\label{tab:factorial_ablation}
\begin{tabular}{lrr}
\toprule
2-hop budget & Visual edges & Accuracy (\%) \\
\midrule
3 (default) & On & 67.1 \\
0 & On & 67.9 \\
3 & Off & 68.6 \\
0 & Off & 68.9 \\
\bottomrule
\end{tabular}
\end{table}

Removing 2-hop expansion changes accuracy by +0.8 pp, removing visual edges by +1.5 pp, and removing both by +1.8 pp. The difference-in-differences is $68.9-68.6-67.9+67.1=-0.5$ pp. These observed differences describe non-additivity.

\subsection{GPT-5 context protocol}
\label{app:gpt5_and_visual_controls}

The two MAPG-augmented GPT-5 conditions in Table~\ref{tab:model_comparison} cover all 1,042 MicroVQA questions using the same prompt format as the open-source runs. Shared context directly reuses the retrieved context supplied to Qwen3-VL-8B Two-Pass, yielding 71.5\%. Own extraction uses GPT-5 for Round-1 entity extraction before retrieval from the same MAPG, yielding 73.3\%. The 1.8-point difference compares retrieval conditioned on different query extractors, with the answer model held fixed.

\subsection{Text-only and knowledge-format controls}
\label{app:knowledge_controls}

Table~\ref{tab:visual_controls} reports the completed Qwen3-VL-8B Two-Pass controls on MicroVQA. Text-only KG removes visual entities and \texttt{visually\_similar} edges while retaining the text pathway. The flat-paragraph condition retains the same retrieved entities, definitions, and relations as full MAPG and changes only their presentation.

\begin{table}[ht]
\centering
\caption{MicroVQA knowledge controls (8B, Two-Pass).}
\label{tab:visual_controls}
\begin{tabular}{lrr}
\toprule
Condition & Accuracy (\%) & $\Delta$ (pp) \\
\midrule
Full MAPG, structured blocks & 67.1 & 0.0 \\
Text-only KG & 66.0 & $-$1.1 \\
Full MAPG, flat paragraph & 66.9 & $-$0.2 \\
\bottomrule
\end{tabular}
\end{table}

On MicroBench, full MAPG and text-only KG reach 58.54\% and 58.15\%, respectively, a 0.39-point difference under the same 8B Two-Pass configuration. These results indicate modest visual-pathway contributions in the evaluated settings. The joint text-only change on MicroVQA ($-1.1$ pp) differs from the sum of the individual visual-entity and visual-edge changes ($-1.3+1.5=+0.2$ pp; Table~\ref{tab:ablation}). This comparison varies different factors from Table~\ref{tab:factorial_ablation}.

The 0.2-point flat-format difference holds knowledge content fixed and supports a limited sensitivity to presentation format in this setting.

% \clearpage

% \input{checklist}

\end{document}